\documentclass{article}
\usepackage{iclr2027_conference,times}

\usepackage{hyperref}
\usepackage{url}
\usepackage{amsmath}
\usepackage{amssymb}
\usepackage{graphicx}
\usepackage{booktabs}
\usepackage{multirow}

\title{AGM: Achievement-Grounded Memory for\\
Closed-Loop Agents with Frozen VLA Policies}

\author{%
Hongbo Gao\textsuperscript{1,*}\enspace
Zeyu Ni\textsuperscript{2,*}\enspace
Xin Wen\textsuperscript{1}\enspace
Siyu Xu\textsuperscript{2}\enspace
Ruifeng Li\textsuperscript{1}\enspace
Rui Xu\textsuperscript{3}\enspace
Yunke Wang\textsuperscript{2,\ensuremath{\dagger}}\\[4pt]
{\normalfont\small
\textsuperscript{1}Harbin Institute of Technology\quad
\textsuperscript{2}University of Sydney\quad
\textsuperscript{3}Wuhan University}\\[2pt]
{\normalfont\small
\textsuperscript{*}Equal contribution\quad
\textsuperscript{\ensuremath{\dagger}}Corresponding author}
}

\hypersetup{
  pdftitle={AGM: Achievement-Grounded Memory for Closed-Loop Agents with Frozen VLA Policies},
  pdfauthor={Hongbo Gao, Zeyu Ni, Xin Wen, Siyu Xu, Ruifeng Li, Rui Xu, Yunke Wang},
  pdfsubject={Public preprint},
  pdfkeywords={vision-language-action models, robot manipulation, task memory, achievement verification}
}

\iclrfinalcopy


\begin{document}
\maketitle

\fancyhead{}
\lhead{\small AGM: Achievement-Grounded Memory}
\rhead{\small Preprint}
\renewcommand{\headrulewidth}{0.4pt}

\begin{abstract}
Frozen vision-language-action (VLA) policies offer broad manipulation
skills but execute open-loop action chunks without tracking task progress,
so the agent cannot reliably decide whether to continue, retry, or
terminate. External memory is a natural remedy, yet we find it can be
harmful when attempted actions are recorded as completed progress:
transient execution failures become persistent task-state errors, and such
a memory can underperform no progress memory at all. We propose
Achievement-Grounded Memory (AGM), a lightweight closed-loop framework for
frozen VLA policies. AGM represents a task as a static subgoal sequence
with a dynamic progress pointer and advances the pointer on physically
verified achievement rather than on attempts. Proprioceptive gripper-load
cues decide when to verify; coherent point tracking verifies grasps, and
language-conditioned cross-view comparison, read by a single trained
2.43M-parameter verification head, verifies placements. The policy,
tracker, and encoder remain frozen, the head is the only trained
component, and deployment needs no auxiliary vision-language model. On the
RoboMME Counting benchmark, AGM reaches $100.0\%$ on PickXTimes and
$84.0\%$ on BinFill, surpassing the strongest memory-augmented baseline by
$7.7$ points on the four-task average, and the gains carry over to a
physical robot, where AGM reaches $100.0\%$ and $82.0\%$. These results
suggest that reliable embodied memory depends more on disciplined state
updates than on memory capacity.
\end{abstract}

\section{Introduction}
\label{sec:intro}

Vision-language-action (VLA) models map visual observations and language
instructions directly to continuous
actions~\citep{brohan2023rt,kim2024openvla,team2024octo,intelligence2025pi_}.
Since adapting such a model to every task is costly, a practical
alternative keeps the policy frozen and composes its skills through
language-conditioned subgoals~\citep{ahn2022can,huang2022inner}.

This strategy exposes a fundamental limitation: task progress is a hidden
state that a frozen policy neither represents nor can recover from its
current observation. In repetitive and counting-based
manipulation~\citep{dai2026robomme}, task stages are perceptually aliased;
after placing an object into a container, the image alone cannot reveal
which repetition is underway. Even when every grasp and placement lies
within the policy's capability, reliable completion requires a persistent
task state that decides when to continue, retry, roll back, or terminate.

External memory appears to be a direct solution~\citep{sridhar2025memer},
yet we find that, under the same frozen policy and subgoal interface, a
progress memory that advances after every attempt can perform \emph{worse}
than no progress memory at all. An attempt is not an achievement: a grasp
may miss, an object may slip during lifting, and a release may fail to
establish the instructed object--target relation, and once such an attempt
is written as progress, a transient failure becomes a persistent task-state
error. The central question thus shifts from \emph{what to store} to
\emph{when to write}: an ordered subgoal sequence with a progress pointer
is often sufficient memory, provided each update rests on observable
execution evidence rather than on the mere occurrence of an attempt.

We instantiate this principle in \textbf{Achievement-Grounded Memory
(AGM)}, a lightweight closed-loop framework for frozen VLA policies. A
gripper-load state machine localizes \emph{when} progress may have changed,
and visual verification decides \emph{whether} the intended physical
transition occurred~\citep{du2023vision,liu2023reflect}. The two achievement
types that anchor repetitive manipulation have different observable
signatures, and AGM sources each from the frozen foundation model best
suited to it: grasp success is a kinematic question of whether the object
rises coherently with the gripper, which a frozen point tracker
\citep[CoTracker3;][]{karaev2024cotracker} answers without any
training~\citep{vecerik2023robotap}; placement success is a semantic
question of whether the instructed object--target relation holds, which AGM
answers by contrasting event-relative front and wrist observations under
frozen SigLIP embeddings~\citep{zhai2023sigmoid} conditioned on the subgoal
language. The verified outcome drives a recoverability-aware pointer
transition: the pointer advances on success, is retained after a failed
grasp, and rolls back after a failed recoverable placement. The VLA policy,
point tracker, and encoder all remain frozen; only a 2.43M-parameter
verification head is trained, and deployment needs no auxiliary
autoregressive vision-language model.

On the RoboMME Counting benchmark~\citep{dai2026robomme}, AGM attains the
best average success rate ($56.56\%$), surpassing the strongest
memory-augmented baseline~\citep{sridhar2025memer} by $7.7$ points and
nearly doubling the frozen base policy~\citep{intelligence2025pi_}. On
subgoals anchored to gripper interactions it reaches $100.00\%$ on
PickXTimes and $84.00\%$ on BinFill, new state-of-the-art results obtained
with 228.6M frozen external parameters and a 2.43M trainable head instead
of a fine-tuned 4B vision-language model. On other subgoals AGM defers to
the frozen policy without degrading it, and controlled comparisons confirm
that the gains arise from verified memory writing rather than from memory
capacity or model scale. On a physical robot, with the policies and
verification head retrained for the platform under an unchanged framework,
AGM yields $100.0\%$ and $82.0\%$ on the two tasks, so the benefit of
achievement-grounded writing is not an artifact of simulation.

Our contributions are summarized as follows:
\begin{itemize}
    \item We identify a failure mode of memory-augmented frozen VLA agents:
    attempt-based progress updates turn transient manipulation failures
    into persistent task-state errors and can underperform no progress
    memory at all.
    \item We propose AGM, which grounds every update of a compact progress
    pointer in observable execution evidence from frozen foundation models
    and supports advancement, retention, and recovery-aware rollback.
    \item We set new state-of-the-art results on PickXTimes and BinFill in
    RoboMME Counting with the best average success rate while training
    only a 2.43M-parameter verification head, and validate the framework
    on a physical robot.
\end{itemize}

\section{Related Work}
\label{sec:related}

\paragraph{Vision-language-action models and skill composition.}
Large-scale VLA models map visual observations and language instructions
directly to low-level
actions~\citep{brohan2022rt,brohan2023rt,team2024octo,kim2024openvla,black2024pi0,intelligence2025pi_},
increasingly trained on large cross-embodiment corpora~\citep{o2024open}.
Since per-task finetuning is costly, a complementary line composes
pretrained skills through language via affordance-grounded
planning~\citep{ahn2022can}, embodied multimodal
planning~\citep{driess2023palm}, policy-code generation~\citep{liang2023code},
and feedback-augmented planning~\citep{huang2022inner}. These systems select
the next skill and assume its completion is observable or handled by
replanning, whereas in repetitive manipulation task stages are perceptually
aliased and subgoal completion cannot be read from the current
observation. AGM leaves the frozen policy and subgoal interface untouched
and supplies this missing task-progress state.

\paragraph{Memory for long-horizon embodied agents.}
Maintaining task state under partial observability is a classical
problem~\citep{kaelbling1998planning}. LLM-based agents address it with
memory streams~\citep{park2023generative}, episodic
reflections~\citep{shinn2023reflexion}, and growing skill
libraries~\citep{wang2023voyager}; in robot learning, SAM2Act adds a
spatial memory architecture to a manipulation policy~\citep{fang2025sam2act},
RoboMME shows that generalist policies fail on non-Markovian manipulation
without external memory~\citep{dai2026robomme}, and MemER retrieves
task-relevant keyframes from past experience with a finetuned
vision-language model~\citep{sridhar2025memer}. These approaches scale
\emph{what} is stored or retrieved, whereas AGM addresses \emph{when} a
write is justified: for repetitive manipulation a compact subgoal pointer
is a sufficient task state, and unverified writes, not limited capacity,
are the dominant failure source.

\paragraph{Verifying manipulation outcomes.}
Execution outcomes have been verified with finetuned vision-language
success detectors~\citep{du2023vision}, multimodal failure
explanation~\citep{liu2023reflect,duan2024aha}, plan-execution
misalignment detection and runtime monitors~\citep{guo2023doremi,zhou2025code},
and environment feedback injected into planning
loops~\citep{huang2022inner}, typically by querying a large model at fixed
intervals or analyzing failures post hoc. AGM instead triggers verification
only at proprioceptive gripper-load transitions and grounds the decision in
frozen perception. Point tracking, matured from dedicated
benchmarks~\citep{doersch2022tap} to robust general-purpose
trackers~\citep{doersch2023tapir,karaev2024cotracker} and used as a
representation for imitation and policy
learning~\citep{vecerik2023robotap,wen2023any}, is repurposed here as
zero-training kinematic evidence of grasp success, while a frozen
vision-language encoder~\citep{zhai2023sigmoid} supplies
language-conditioned placement evidence.

\begin{figure}[t]
    \centering
    \includegraphics[width=\textwidth]{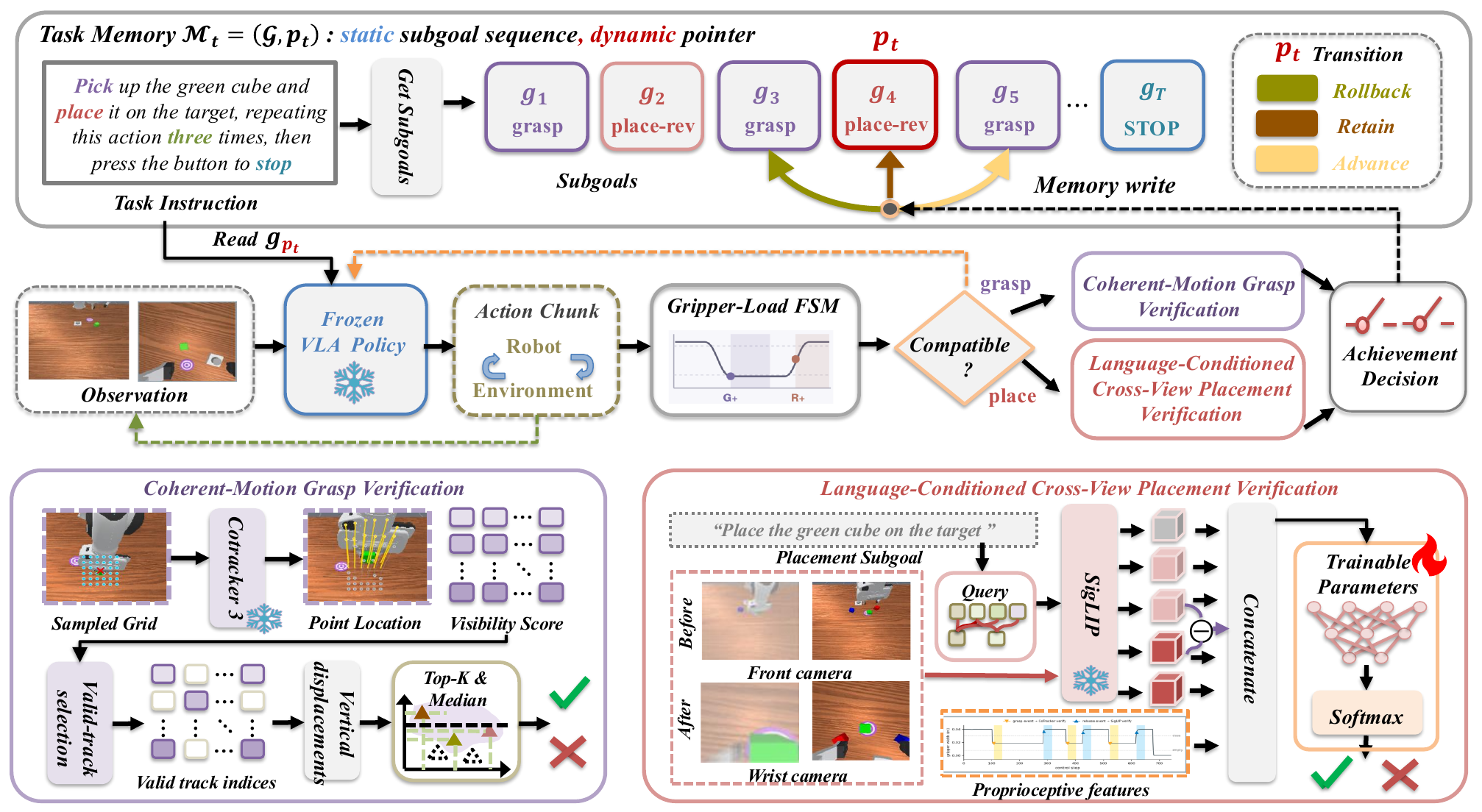}
    \caption{
\textbf{Overview of AGM.}
\emph{Top:} the instruction expands into a typed subgoal sequence
$\mathcal{G}$, which with a progress pointer forms the task memory
$\mathcal{M}_t=(\mathcal{G},p_t)$. The frozen VLA acts on $g_{p_t}$; a
gripper-load state machine detects interaction events, and only events
compatible with the current subgoal type trigger verification, since an
interaction alone is not evidence of achievement. The verdict updates the
pointer by Eq.~\eqref{eq:memory_transition}; the colored arrows depict
that rule, not three outcomes reachable at one subgoal.
\emph{Bottom:} grasp achievement is verified by coherent upward point
motion under a frozen tracker, placement by language-conditioned
pre/post-release cross-view evidence under a frozen encoder. Snowflakes
mark frozen modules; only the 2.43M verification head (flame) is trained.
    }
    \label{fig:overview}
\end{figure}

\section{Method}
\label{sec:method}

\subsection{Problem Formulation and Overview}
\label{sec:method_overview}

We consider a frozen, subgoal-conditioned VLA policy
$\pi_{\theta}(A_t\mid o_t,g_t)$ with $\theta$ fixed throughout training and
evaluation; $o_t$ contains front- and wrist-view images, gripper opening
width, and end-effector pose, $g_t$ is a language subgoal, and $A_t$ an
action chunk. Figure~\ref{fig:overview} summarizes the framework.

For repetitive manipulation, the task instruction is deterministically expanded
into an ordered sequence
\begin{equation}
\mathcal{G}
=
\{(g_k,c_k)\}_{k=1}^{T},
\qquad
c_k
\in
\mathcal{C}
=
\{
\texttt{grasp},
\texttt{place-rev},
\texttt{place-irrev},
\texttt{other}
\},
\end{equation}
where $g_k$ is provided to the frozen VLA and $c_k$ denotes the subgoal
type. Planar placement is treated as recoverable through re-grasping,
container placement as non-recoverable after release. The type labels
determine the compatible verification event and recovery behavior and are
not policy outputs. In RoboMME Counting we use the benchmark-provided
decomposition, cross-checked with an instruction-only parser that accesses
neither environment state nor task ground truth.

AGM maintains a compact task memory
\begin{equation}
\mathcal{M}_t=(\mathcal{G},p_t),
\qquad
p_t\in\{1,\ldots,T\},
\qquad
A_t
\sim
\pi_{\theta}
\left(
\cdot\mid o_t,g_{p_t}
\right),
\end{equation}
where $\mathcal{G}$ remains fixed within an episode, $p_t$ is the only
persistent task-progress state updated online, and the frozen policy is
conditioned on the indexed subgoal $g_{p_t}$. The pointer retains
task-stage information that the current observation cannot supply,
disambiguating visually similar repetitions without storing the full
interaction history.

AGM updates the pointer only after compatible interaction events. Let
\begin{equation}
\mathcal{E}_{\mathrm{ver}}
=
\left\{
t
\mid
\chi(e_t,c_{p_t})=1
\right\}
=
\{\tau_1,\tau_2,\ldots\},
\label{eq:verification_events}
\end{equation}
where $e_t$ is the detected event and $\chi$ retains only event--subgoal pairs
supported by AGM. After an event $\tau_n$, AGM collects an event-dependent
observation window (the subsequent lifting motion for grasping, a
post-release settling period for placement) and verifies at
$\bar{\tau}_n$. The resulting achievement decision
$\widehat{v}_n\in\{0,1\}$ determines whether the pointer advances, stays,
or rolls back. Under nominal execution the pointer changes only when such a
decision becomes available; the terminal consistency check and bounded
deadlock safeguards are described below.

\subsection{Event-Triggered Achievement Verification}
\label{sec:achievement_verification}

\subsubsection{Interaction Event Localization}
\label{sec:event_localization}

Let $w_t$ denote the gripper opening width. A hysteretic finite-state machine
produces
\begin{equation}
(\xi_t,e_t)
=
F_{\mathrm{fsm}}(\xi_{t-1},w_t),
\qquad
e_t
\in
\left\{
\varnothing,
\mathsf{G}^{+},
\mathsf{R}^{+},
\mathsf{R}^{0}
\right\},
\label{eq:fsm}
\end{equation}
where $\xi_t$ is the internal load state. Here $\mathsf{G}^{+}$ denotes a
stable loaded grasp, $\mathsf{R}^{+}$ release from a loaded state, and
$\mathsf{R}^{0}$ release from an empty state. The compatibility function in
Eq.~\eqref{eq:verification_events} satisfies $\chi(\mathsf{G}^{+},c)=1$
only for grasp subgoals and $\chi(\mathsf{R}^{+},c)=1$ only for placement
subgoals; empty releases and incompatible pairs are ignored. These events
localize potential state transitions but are not themselves evidence of
success.

\subsubsection{Coherent-Motion Grasp Verification}
\label{sec:grasp_verification}

A successful grasp moves the object upward with the gripper during
lifting. When $e_{\tau_n}=\mathsf{G}^{+}$, AGM records the end-effector
position $\mathbf{x}^{ee}_{\tau_n}$, buffers the subsequent front-view
lifting clip, and projects the recorded position into the image with camera
extrinsics $E$ and intrinsics $K$:
\begin{equation}
\mathbf{x}_{c}
=
E
\begin{bmatrix}
\mathbf{x}^{ee}_{\tau_n}\\
1
\end{bmatrix},
\qquad
\widetilde{\mathbf{u}}
=
K\mathbf{x}_{c},
\qquad
\mathbf{u}_{n}
=
\left(
\frac{\widetilde{u}_1}{\widetilde{u}_3},
\frac{\widetilde{u}_2}{\widetilde{u}_3}
\right).
\label{eq:grasp_projection}
\end{equation}
The end-effector pose only localizes the query region and determines when
sufficient lifting motion has been observed.

Let $\mathcal{P}_{n}^{0}=\{\mathbf{u}_{n,i}^{0}\}_{i=1}^{N_q}$ be a regular
grid around $\mathbf{u}_{n}$ and
$\mathbf{I}_{n}^{f}=\{I_{n,\ell}^{f}\}_{\ell=1}^{L}$ the $L$-frame lifting
clip. A frozen point tracker, instantiated with
CoTracker3~\citep{karaev2024cotracker}, returns trajectories and visibility
scores:
\begin{equation}
(\mathbf{U}_{n},\mathbf{V}_{n})
=
\mathcal{F}_{\phi}
\left(
\mathbf{I}_{n}^{f},
\mathcal{P}_{n}^{0}
\right),
\qquad
\phi\ \text{fixed},
\label{eq:frozen_point_tracker}
\end{equation}
where $\mathbf{U}_{n,i,\ell}\in\mathbb{R}^{2}$ is the image location of
query point $i$ at frame $\ell\in\{1,\ldots,L\}$ and $V_{n,i,\ell}\in[0,1]$
its visibility.

The valid track indices and their vertical displacements are
\begin{equation}
\mathcal{I}_{n}^{\mathrm{vis}}
=
\left\{
i
\mid
V_{n,i,L}
\geq
\eta_{\mathrm{vis}}
\right\},
\qquad
d_{n,i}
=
U_{n,i,1}^{(y)}
-
U_{n,i,L}^{(y)},
\quad
i\in\mathcal{I}_{n}^{\mathrm{vis}}.
\label{eq:vertical_displacement}
\end{equation}
Since the image $y$-axis points downward, $d_{n,i}>0$ indicates upward
motion.

Since the query region may contain background points, AGM retains the
fraction $\rho$ of valid tracks with the largest upward displacement and
computes
\begin{equation}
r_{n}^{\mathrm{G}}
=
\operatorname{Median}
\left(
\operatorname{Top}_{\rho}
\left\{
d_{n,i}
\mid
i\in\mathcal{I}_{n}^{\mathrm{vis}}
\right\}
\right).
\label{eq:grasp_motion_score}
\end{equation}
The high-response selection suppresses static background tracks and median
aggregation limits the effect of isolated tracking errors, so grasp
completion is decided by the observed motion of the object rather than the
commanded motion of the gripper.

\subsubsection{Language-Conditioned Cross-View Placement Verification}
\label{sec:placement_verification}

A loaded release does not guarantee the desired object--target relation.
For each placement subgoal AGM derives a normalized achievement query
$q_k=\mathcal{Q}(g_k)$, which states the required relation without
repetition indices.

When $e_{\tau_n}=\mathsf{R}^{+}$, AGM extracts event-relative windows before
and after release from the front ($f$) and wrist ($w$) cameras, aggregates
them temporally into $\overline{I}_{v,n}^{-}$ and $\overline{I}_{v,n}^{+}$
for $v\in\{f,w\}$, and encodes them with a frozen vision-language encoder:
\begin{equation}
z_{v,n}^{-}
=
\operatorname{Enc}_{\omega}^{\mathrm{img}}
\left(
\overline{I}_{v,n}^{-}
\right),
\quad
z_{v,n}^{+}
=
\operatorname{Enc}_{\omega}^{\mathrm{img}}
\left(
\overline{I}_{v,n}^{+}
\right),
\quad
z_{q,n}
=
\operatorname{Enc}_{\omega}^{\mathrm{text}}
\left(
q_{p_{\tau_n}}
\right),
\quad
\omega\ \text{fixed}.
\end{equation}

The event representation combines cross-view appearance, event-induced change,
proprioceptive context, and the requested object--target relation:
\begin{equation}
\mathbf{f}_n
=
\left[
z_{f,n}^{-};
z_{f,n}^{+};
z_{f,n}^{+}-z_{f,n}^{-};
z_{w,n}^{-};
z_{w,n}^{+};
\mathbf{z}_{n}^{\mathrm{prop}};
z_{q,n}
\right],
\label{eq:event_feature}
\end{equation}
where $\mathbf{z}_{n}^{\mathrm{prop}}$ summarizes the proprioceptive window
around the event; the front-view difference captures the dominant state
change and the wrist view adds near-field evidence.

A lightweight multilayer perceptron $h_{\psi}$ (2.43M trainable
parameters; the feature encoders remain frozen) predicts the placement
achievement probability:
\begin{equation}
s_{n}^{\mathrm{P}}
=
\left[
\operatorname{Softmax}
\left(
h_{\psi}(\mathbf{f}_n)
\right)
\right]_{\mathrm{achieved}}.
\label{eq:placement_score}
\end{equation}

\subsection{Achievement-Grounded Memory Transitions}
\label{memory_transition}

The two branches produce different evidence and are thresholded
separately:
\begin{equation}
\widehat{v}_n
=
\begin{cases}
\mathbb{I}
\left[
r_{n}^{\mathrm{G}}
\geq
\delta_{\mathrm{G}}
\right],
& e_{\tau_n}=\mathsf{G}^{+},\\
\mathbb{I}
\left[
s_{n}^{\mathrm{P}}
\geq
\kappa_{c_{p_{\bar{\tau}_n}}}
\right],
& e_{\tau_n}=\mathsf{R}^{+},
\end{cases}
\label{eq:achievement_decision}
\end{equation}
with coherent-lift threshold $\delta_{\mathrm{G}}$ and
placement-type-specific acceptance threshold $\kappa_c$.

Let $p_n^{-}=p_{\bar{\tau}_n}$ and $p_n^{+}$ denote the pointer before and
after the $n$-th memory write, $\zeta_k\in\{0,1\}$ indicate whether subgoal
$k$ admits rollback, and $\mathcal{B}(p)$ return its recovery subgoal. The
nominal transition is
\begin{equation}
p_n^{+}
=
\begin{cases}
\min(p_n^{-}+1,T),
& \widehat{v}_n=1,\\
\mathcal{B}(p_n^{-}),
& \widehat{v}_n=0
\ \land\
\zeta_{p_n^{-}}=1,\\
p_n^{-},
& \text{otherwise}.
\end{cases}
\label{eq:memory_transition}
\end{equation}
and the updated memory is
$\mathcal{M}_{\bar{\tau}_n}^{+}=(\mathcal{G},p_n^{+})$. A verified
achievement advances to the next subgoal, a rejected grasp retains the
current state, and a failed recoverable placement returns to its associated
grasp subgoal so that the grasp--place operation is repeated as a whole;
memory writing thus supports advancement, retention, and rollback rather
than attempt counting.

Container placement is non-recoverable after release: a false rejection may
insert an additional object and invalidate the count, so AGM uses a more
permissive acceptance threshold and bounds consecutive rejections for these
events, whereas recoverable planar placement uses a stricter threshold
because a failure can be corrected by re-grasping. Before the final
termination action, AGM rechecks the most recent recoverable placement with
a visual occupancy test independent of the learned head; a failed recheck
returns the pointer to the associated grasp subgoal.

Achievement-grounded updates apply only to subgoals admitted by $\chi$;
other subgoals are delegated to the frozen base policy, and bounded timeout
and retry rules act only when evidence remains unavailable or repeated
rejection would otherwise deadlock the controller.

\subsection{Deployment-Aligned Verifier Training}
\label{sec:verifier_training}

The verification head is trained offline on 788 execution events collected
on the official RoboMME training split, with seeds disjoint from
evaluation. Events come from two complementary controllers: AGM itself,
covering the deployment distribution, and an attempt-based controller that
advances without verification and supplies the missed grasps, misplaced
objects, and incorrect progressions scarce in planner-generated
demonstrations. Environment predicates serve only to
construct labels and are unavailable during evaluation: object lift for
grasps, target-region containment for planar placements, and
container-count increase for container placements.

Grasp and release events share the feature map of
Eq.~\eqref{eq:event_feature}, with grasp windows bracketing the lifting
motion; grasp samples provide auxiliary supervision only, since deployment
decides grasp completion exclusively by the point-tracking test. For irreversible placement, the
post-release window is sampled at exactly the deployment verification time,
since even a 20-frame mismatch substantially degrades downstream
performance (Appendix~\ref{sec:supp-window}). The head is trained with
class-weighted cross-entropy against label imbalance; only its 2.43M
parameters are updated, and the policy, encoder, and point tracker remain
frozen.

\section{Experiments}
\label{sec:experiments}

Our experiments ask whether achievement-grounded memory improves task
success over frozen execution and existing memory-augmented approaches
(Q1), whether the gains stem from task-progress tracking rather than
low-level skill (Q2), how much the visual grasp evidence and the
reversibility-aware thresholds each contribute (Q3), and whether the same
gains hold on a physical robot (Q4).

\subsection{Experimental Setup}
\label{sec:exp_setup}

For \textbf{simulation}, we evaluate on the four RoboMME Counting
tasks~\citep{dai2026robomme} (PickXTimes, BinFill, SwingXTimes, StopCube),
reporting success rate on the official 50-episode test split with seeds
disjoint from training; an episode succeeds only if the robot presses the
stop button after exactly the instructed count within 1{,}300 steps.
Baselines are the frozen $\pi_{0.5}$, $\pi_{0.5}$ with past-action context,
SAM2Act+~\citep{fang2025sam2act}, SimpleSG with a LoRA-finetuned Qwen3-VL-4B
or prompted Gemini-2.5-Pro, and MemER~\citep{sridhar2025memer}, all under
the official evaluation. Our pipeline reproduces the released
$\pi_{0.5}$~\citep{intelligence2025pi_} at $42.0$/$30.0$ on
PickXTimes/BinFill, close to the official $42.89$/$30.00$; AGM entries
average two independent runs. AGM adds to the frozen 3.35B backbone a
frozen CoTracker3 tracker (25.4M)~\citep{karaev2024cotracker}, a frozen
SigLIP-base encoder (203.2M)~\citep{zhai2023sigmoid}, and the 2.43M
verification head, its only trained component; thresholds, state machine,
and transition rules are shared across all tasks and episodes. For the
cross-backbone study we train a subgoal-conditioned
$\pi_0$~\citep{black2024pi0} with LoRA on the official demonstrations, as no
official checkpoint exists for this suite (details in
Appendices~\ref{sec:supp-method}--\ref{sec:supp-sim}).
On the \textbf{real robot}, we deploy AGM for PickXTimes and BinFill with
10 trials per count $N{=}1,\ldots,5$ (50 per task) under the same success
criterion. Following the benchmark configuration, the bare policies are
retrained on real demonstrations conditioned on the task instruction alone,
their AGM counterparts with subgoal conditioning under a matched data
budget, and one verification head shared by both backbones is retrained on
real teleoperated events (Appendix~\ref{sec:supp-real}).

\begin{figure}[t]
    \centering
    \includegraphics[width=\linewidth]{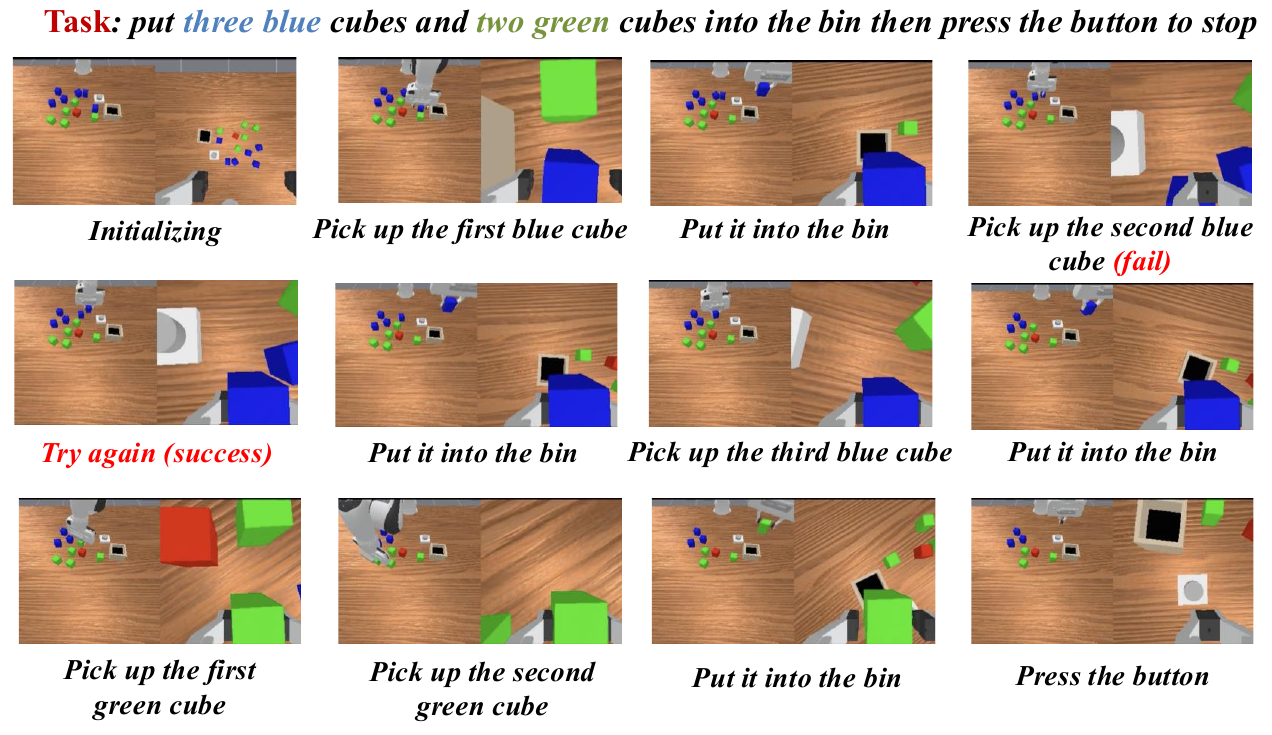}
    \caption{
    \textbf{Error correction in simulation.} The instruction requires three
    blue and two green cubes in the bin. The second grasp fails: verification
    rejects it, the pointer is retained, and the policy retries
    successfully, whereas an attempt-based memory would have recorded the
    failure as progress and finished one cube short.
    }
    \label{fig:sim_rollout}
\end{figure}

\subsection{Main Results (Q1)}
\label{sec:exp_main}

\begin{table}[t]
\centering
\caption{
Success rates (\%) on the RoboMME Counting test split (50 episodes per
task; AGM averaged over two runs). \emph{External}: frozen parameters added
beyond the 3.35B $\pi_{0.5}$ backbone; \emph{Trainable}: parameters updated
per method. $^{\dagger}$: subgoals not anchored to gripper interactions,
where AGM follows the frozen policy. $^{\ddagger}$: independent policy not
built on $\pi_{0.5}$. API: proprietary model of undisclosed size;
\emph{n/r}: not reported. Best per column in \textbf{bold}.
}
\label{tab:main}
\footnotesize
\setlength{\tabcolsep}{1pt}
\begin{tabular*}{\textwidth}{
@{\extracolsep{\fill}}
l
ccccc
cc
@{}
}
\toprule
& \multicolumn{5}{c}{Success Rate (\%)}
& \multicolumn{2}{c}{Params.} \\
\cmidrule(lr){2-6}
\cmidrule(lr){7-8}
Method
& PickXTimes
& BinFill
& SwingXTimes
& StopCube
& Avg.
& External
& Trainable \\
\midrule
Frozen $\pi_{0.5}$
& 42.89 & 30.00 & 35.56 & \textbf{6.67} & 28.78
& 0 & 0 \\
$\pi_{0.5}$ w/ past actions
& 58.33 & 26.67 & 26.67 & 4.67 & 29.09
& 0 & $\approx$3.35B \\
SAM2Act+$^{\ddagger}$
& 76.00 & 40.00 & 25.33 & 0.00 & 35.33
& -- & \emph{n/r} \\
SimpleSG + Qwen3-VL-4B
& 95.30 & 77.60 & 5.10 & 0.40 & 44.60
& 4B & LoRA (\emph{n/r}) \\
SimpleSG + Gemini-2.5-Pro
& 63.00 & 46.00 & 45.00 & 2.00 & 39.00
& API & 0 \\
MemER
& 79.33 & 56.67 & \textbf{59.33} & 0.00 & 48.83
& 4B & LoRA (\emph{n/r}) \\
\midrule
\textbf{AGM (Ours)}
& \textbf{100.00} & \textbf{84.00} & 35.56$^{\dagger}$ & \textbf{6.67}$^{\dagger}$ & \textbf{56.56}
& 228.6M (frozen) & \textbf{2.43M} \\
\bottomrule
\end{tabular*}
\end{table}

\begin{table}[t]
\centering
\caption{
Success rates (\%) by instructed repetition count $N$. \emph{Sim} rows use
the official test split (16/10/12/6/6 and 10/10/16/8/6 episodes per $N$
for PickXTimes and BinFill; episode-weighted means match
Table~\ref{tab:main}); $\pi_{0.5}$ is our reproduction of the official
checkpoint and $\pi_0$ a subgoal-conditioned LoRA, marked $^{\dagger}$ since
full instructions are out-of-distribution for it. \emph{Real} rows use 10
trials per count with policies and verification head retrained for the
platform under an unchanged framework. Best per column in \textbf{bold}.
}
\label{tab:perN}
\footnotesize
\setlength{\tabcolsep}{4.5pt}
\begin{tabular*}{\textwidth}{@{\extracolsep{\fill}}ll ccccc ccccc@{}}
\toprule
& & \multicolumn{5}{c}{PickXTimes} & \multicolumn{5}{c}{BinFill} \\
\cmidrule(lr){3-7}\cmidrule(lr){8-12}
& Method & $N{=}1$ & $N{=}2$ & $N{=}3$ & $N{=}4$ & $N{=}5$
         & $N{=}1$ & $N{=}2$ & $N{=}3$ & $N{=}4$ & $N{=}5$ \\
\midrule
\multirow{4}{*}{Sim}
& $\pi_0^\dagger$   & 12.5 & 0.0 & 0.0 & 0.0 & 0.0
                    & 0.0 & 0.0 & 0.0 & 0.0 & 0.0 \\
& $\pi_0$+AGM       & 50.0 & 40.0 & 33.3 & 50.0 & 50.0
                    & 30.0 & 10.0 & 0.0 & 0.0 & 0.0 \\
& $\pi_{0.5}$       & \textbf{100.0} & 30.0 & 16.7 & 0.0 & 0.0
                    & 90.0 & 20.0 & 18.8 & 12.5 & 0.0 \\
& $\pi_{0.5}$+AGM   & \textbf{100.0} & \textbf{100.0} & \textbf{100.0} & \textbf{100.0} & \textbf{100.0}
                    & \textbf{100.0} & \textbf{95.0} & \textbf{84.4} & \textbf{62.5} & \textbf{66.7} \\
\midrule
\multirow{4}{*}{Real}
& $\pi_0$    & 20.0 & 0.0 & 0.0 & 0.0 & 0.0 & 0.0 & 0.0 & 0.0 & 0.0 & 0.0 \\
& $\pi_0$+AGM       & 60.0 & 50.0 & 30.0 & 60.0 & 50.0 & 30.0 & 20.0 & 0.0 & 0.0 & 0.0 \\
& $\pi_{0.5}$       & \textbf{100.0} & 20.0 & 10.0 & 0.0 & 0.0 & \textbf{100.0} & 10.0 & 0.0 & 0.0 & 0.0 \\
& $\pi_{0.5}$+AGM   & \textbf{100.0} & \textbf{100.0} & \textbf{100.0} & \textbf{100.0} & \textbf{100.0} & \textbf{100.0} & \textbf{100.0} & \textbf{80.0} & \textbf{70.0} & \textbf{60.0} \\
\bottomrule
\end{tabular*}
\end{table}

Table~\ref{tab:main} summarizes task success and parameterization. AGM
attains the best four-task average ($56.56\%$), surpassing the strongest
baseline MemER ($48.83\%$) by $7.7$ points and nearly doubling the frozen
$\pi_{0.5}$ ($28.78\%$). On the two tasks anchored to gripper
interactions, where perceptual aliasing is most damaging, the gains are
decisive: $100.00\%$ on PickXTimes and $84.00\%$ on BinFill,
both new state-of-the-art results. Past-action conditioning ($29.09\%$)
recovers almost none of the gap, so the missing information is task
progress rather than short-term context. AGM is also far more economical:
SimpleSG and MemER attach a LoRA-finetuned 4B external VLM, whereas AGM
adds 228.6M frozen parameters and trains only the 2.43M head. On
SwingXTimes and StopCube, whose subgoals are not gripper-anchored, AGM
defers to the frozen policy by construction and matches it exactly.

\begin{figure}[t]
    \centering
    \includegraphics[width=\textwidth]{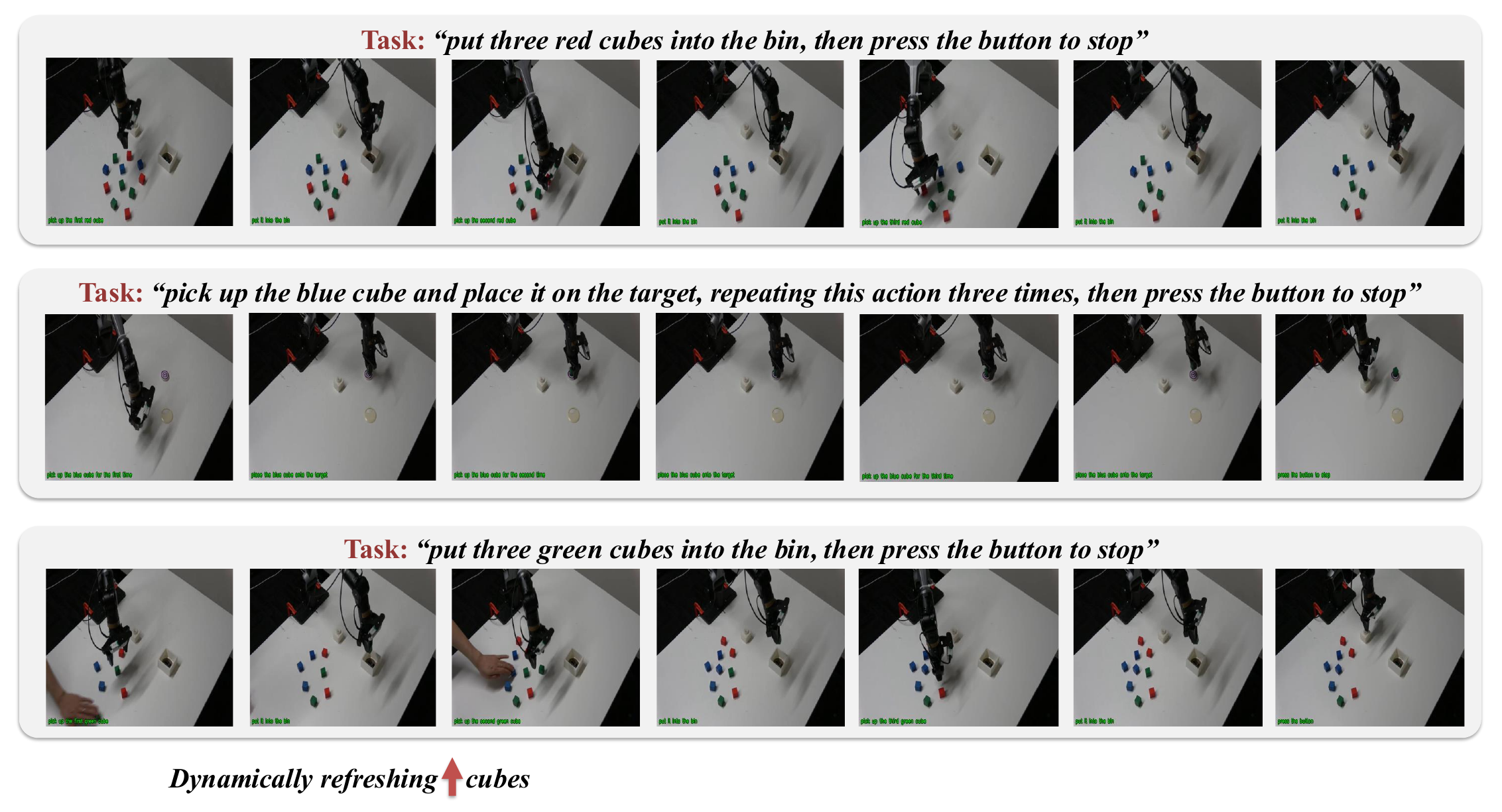}
    \caption{
    \textbf{Real-robot rollouts.} Filling the bin with three red cubes,
    placing a blue cube on the target three times, and filling the bin with
    three green cubes while a human replenishes cubes mid-task; the
    heart-shaped initial layout (top) and the mid-task replenishment
    (bottom) do not appear in the demonstrations. Overlays show the subgoal
    issued at each moment.
    }
    \label{fig:real_rollout}
\end{figure}

\subsection{Effect of Repetition Depth (Q2)}
\label{sec:exp_perN}

Table~\ref{tab:perN} breaks down success by the instructed count $N$. The
frozen $\pi_{0.5}$ is nearly perfect at $N{=}1$ ($100.0$/$90.0$ on
PickXTimes/BinFill) but collapses to $0.0$ on both tasks by $N{=}5$: it
has the required skills and lacks only the task state that decides whether
to continue or stop, so counting errors compound with $N$. With AGM,
PickXTimes stays at $100.0\%$ for every $N$, eliminating progress drift,
while BinFill decays gradually to $66.7\%$ with the remaining failures in
later, physically harder repetitions rather than in progress tracking. The
subgoal-conditioned $\pi_0$ is barely functional alone, yet the identical
configuration, with no threshold or rule retuned, lifts it to
$44.0\%$/$8.0\%$: achievement-grounded writing is a property of the
interface rather than of a particular policy. Figure~\ref{fig:sim_rollout}
shows the mechanism in a complete rollout: a failed grasp is rejected, the
pointer is retained, and the episode recovers without a progress error.

\subsection{Ablations (Q3)}
\label{sec:exp_ablation}

\begin{table}[t]
\centering
\caption{
Ablations on grasp-verification modality and reversibility-aware thresholds
(success rate \%, official test split, $n{=}50$ per cell, same frozen
$\pi_{0.5}$ and head as Table~\ref{tab:main}; row 1 is the full method,
BinFill as mean of two runs). \emph{Proprioception only}
replaces the point-tracking grasp check with a gripper-load and
end-effector-lift test; \emph{flattened} removes the
reversibility-dependent threshold switch.
}
\label{tab:ablation}
\footnotesize
\setlength{\tabcolsep}{4pt}
\begin{tabular}{llcc}
\toprule
Grasp verification & Thresholds & PickXTimes & BinFill \\
\midrule
CoTracker (full)    & reversibility-aware & \textbf{100.0} & \textbf{84.0} \\
Proprioception only & reversibility-aware & 98.0           & 78.0 \\
CoTracker           & flattened           & 98.0           & 74.0 \\
\bottomrule
\end{tabular}
\end{table}

Table~\ref{tab:ablation} isolates the two verification design choices with
everything else fixed. Replacing the point-tracking test with a
proprioception-only check (stable gripper load plus end-effector lift)
keeps PickXTimes at $98.0\%$ and reaches $78.0\%$ on BinFill: verified
writing itself carries most of the benefit, and even this vision-free
variant beats the strongest external-VLM baseline, while the
coherent-motion test adds a consistent residual margin by confirming that
the object itself moves with the gripper. Flattening the
reversibility-aware thresholds barely affects PickXTimes ($98.0\%$), whose
leaked failures are absorbed by rollback, but drops BinFill from $84.0\%$
to $74.0\%$: strict acceptance falsely rejects correct irreversible drops,
and with the object already in the container the pointer stalls on a
completed subgoal until the budget is spent, so the value of the case split
concentrates on the irreversible side.

\subsection{Real-Robot Validation (Q4)}
\label{sec:exp_real}

The lower block of Table~\ref{tab:perN} reports physical deployment. Even
retrained on real demonstrations, the bare policies show the same
progress-tracking failure that motivates the method: $\pi_{0.5}$ solves
$N{=}1$ on both tasks, falls to $20.0\%$ and $10.0\%$ at $N{=}2$, and
reaches zero on both by $N{=}4$ although it executes individual grasps and
placements competently; retraining improves skills but cannot supply the
missing task state.

AGM recovers that state: PickXTimes reaches $100.0\%$ at every count, and
BinFill holds $100.0\%$ through $N{=}2$ before decaying to $60.0\%$ at
$N{=}5$. The weaker $\pi_0$ backbone is nearly non-functional alone, yet
the same AGM configuration, with the verification head shared between
backbones, lifts it substantially, so the benefit again follows the
interface rather than the backbone. The residual BinFill gap at large $N$
is physical rather than representational: failures concentrate in late
repetitions, where the increasingly occupied container makes placement
harder, while the pointer remained consistent with the executed
achievements in every trial we inspected.

\paragraph{Generalization and robustness beyond the demonstrations.}
Single-color BinFill demonstrations stop at $N{=}4$, whereas $60\%$ of the
BinFill $N{=}5$ trials instruct a single color, so most of that cell
evaluates AGM's progress tracking on a configuration the checkpoint never
saw; it still reaches $60.0\%$ against $70.0\%$ at the demonstrated
$N{=}4$, and every trial starts from cube layouts, counts, and color
mixtures unseen in training. Robustness is probed by the $40\%$
of BinFill trials that are dynamic: under the benchmark's dynamic-refresh
rule a human adds cubes at arbitrary moments, mid-grasp, mid-release into
the bin, or while the arm waits for a cube of the instructed color.
Because a memory write is considered only at gripper-load events and the
verifier then judges the instructed object--target relation rather than
the whole scene, the changing layout and the hand in view produced no
false acceptance, false rejection, or forced advance in any dynamic trial,
and the policy kept executing the current subgoal until a suitable cube
appeared (Appendix~\ref{sec:supp-eval}). Figure~\ref{fig:real_rollout}
shows both cases: the heart-shaped initial layout (top) and the mid-task
replenishment (bottom) are absent from the demonstrations, yet the verified
count remains correct throughout.

\section{Conclusion}
\label{sec:conclusion}

We presented AGM, a lightweight framework that closes the loop around a
frozen VLA policy by grounding every update of a compact task memory in
observable execution evidence: gripper events decide when to verify, frozen
foundation models supply kinematic and semantic evidence of what was
achieved, and the verdict advances, retains, or rolls back a progress
pointer. Training only a 2.43M-parameter head, AGM sets new
state-of-the-art results on PickXTimes and BinFill with the best four-task
average, transfers across backbones without retuning, and delivers the same
decisive gains on a physical robot under an unchanged framework. Its
verification vocabulary is still anchored to gripper interactions;
extending it beyond prehensile events and widening the physical evaluation
are next steps. More broadly, for long-horizon autonomy the decisive
question is not how much memory an agent has but what evidence entitles it
to write.


\bibliography{references}
\bibliographystyle{iclr2027_conference}

\clearpage
\appendix
%

\section{Implementation Details}
\label{sec:supp-method}

Unless noted otherwise, the values below are the simulation configuration
behind every number in Table~\ref{tab:main} and Table~\ref{tab:ablation}. Real-robot values, where they
differ, appear in Appendix~\ref{sec:supp-real}. No threshold is tuned per task or per
repetition count; one configuration serves all four tasks and every
episode.

\subsection{Subgoal Sequences}
\label{sec:supp-subgoals}

The instruction expands into $T = 2N{+}1$ subgoals, alternating grasp and
placement for $N$ repetitions and closing with the terminal action. The
repetition index is carried by the grasp subgoal, while the placement
subgoal is stated without it. On BinFill the grasp subgoals read
\emph{``pick up the \{ordinal\} \{color\} cube''} and every placement reads
\emph{``put it into the bin''}; on PickXTimes the grasp subgoals read
\emph{``pick up the \{color\} cube for the \{ordinal\} time''} and every
placement reads \emph{``place the \{color\} cube onto the target''}. The
subgoal in force at each pointer transition is visible in the overlays of
Figures~\ref{fig:sim-binfill}--\ref{fig:sim-pickx-red}.

Placement verification is the only branch conditioned on language. The
normalized query $q_k=\mathcal{Q}(g_k)$ of
Section~\ref{sec:placement_verification} maps subgoals differing only in repetition
index onto a common query, so what a placement must evidence is the
object--target relation, not its position in the sequence. Which repetition
is underway is carried by the pointer.

\subsection{Gripper-Load State Machine}
\label{sec:supp-fsm}

The hysteretic state machine of Eq.~\eqref{eq:fsm} consumes the gripper opening width
$w_t$ and emits $e_t \in \{\varnothing, G^+, R^+, R^0\}$. A closure is
entered below $0.030$ and released above $0.035$ in normalized width units,
and a closure narrower than $0.008$ counts as empty. The gap between entry
and exit thresholds suppresses event chatter while the fingers are in
compliant contact, and a load state must hold for five consecutive frames
before it is confirmed. Only confirmed transitions emit events. Of those,
only pairs admitted by the compatibility function $\chi$ of Eq.~\eqref{eq:verification_events} reach
the verifier: $G^+$ with a grasp subgoal, $R^+$ with either placement type.
Empty releases and mismatched pairs are discarded without consuming a
verification.

\subsection{Coherent-Motion Grasp Verification}
\label{sec:supp-grasp}

Eq.~\eqref{eq:grasp_projection} projects the end-effector position at the grasp instant into the
front view, and a regular $6 \times 6$ query grid ($N_q = 36$) is seeded
within a half-width of $22$\,px about that projection, covering a $45$\,px
square.

The lifting clip has no fixed horizon. Frames accumulate from the grasp
event until the end-effector has risen $0.03$\,m, at which point the clip
closes and passes to the tracker, so the length $L$ in Eq.~\eqref{eq:frozen_point_tracker} follows the
executed motion rather than a constant. Table~\ref{tab:cliplen} gives its
empirical distribution.

\begin{table}[htbp]
\centering
\caption{Lifting-clip length $L$ in frames over the reported runs. The clip
closes once a $0.03$\,m lift is observed, so $L$ reflects the executed lift
rather than a fixed horizon.}
\label{tab:cliplen}
\small
\begin{tabular}{@{}lrrrr@{}}
\toprule
Task & Min & Median & Mean & Max \\
\midrule
PickXTimes ($n = 126$) & 7 & 12 & 13.2 & 22 \\
BinFill ($n = 140$)    & 5 & 14 & 15.3 & 30 \\
\bottomrule
\end{tabular}
\end{table}

A track survives when its terminal visibility exceeds
$\eta_{\mathrm{vis}} = 0.5$ (Eq.~\eqref{eq:vertical_displacement}). Among survivors, the fraction
$\rho = 1/3$ with the largest upward displacement is retained, subject to a
floor of six points so the aggregate stays defined when few tracks survive.
Their median is the response $r^G_n$ of Eq.~\eqref{eq:grasp_motion_score}, and the grasp is accepted
at $r^G_n \ge \delta_G = 4$\,px.

\subsection{Language-Conditioned Placement Verification}
\label{sec:supp-placement}

Pre- and post-event windows aggregate $K = 4$ frames from each camera. The
post-release window opens after a settling delay of 30 frames for
recoverable planar placement and 40 frames for irreversible container
placement. These are the offsets at which the verifier's training events are
sampled (Appendix~\ref{sec:supp-events}), so the visual evidence is drawn from the same
post-event moment in training and at deployment; Appendix~\ref{sec:supp-window} reports what
happens when it is not.

The feature map of Eq.~\eqref{eq:event_feature} concatenates five $768$-dimensional visual
embeddings, a $768$-dimensional text embedding, and a proprioceptive summary
of width $2d + 2$, where $d$ is the raw proprioceptive dimension and the two
extra entries are the extrema of the gripper column over the window. With
$d = 8$ in simulation the input width is $4626$.
Table~\ref{tab:head-params} breaks down the resulting parameter count.

\begin{table}[htbp]
\centering
\caption{Parameter count of the verification head, the only trained
component of AGM. Input widths are $4626$ and $4624$ respectively. The
frozen SigLIP encoder (203.2M) and CoTracker3 tracker (25.4M) contribute no
trainable parameters.}
\label{tab:head-params}
\small
\begin{tabular}{@{}lrr@{}}
\toprule
Layer & Simulation & Real robot \\
\midrule
$\mathrm{Linear}(\cdot, 512)$ & $2{,}369{,}024$ & $2{,}368{,}000$ \\
$\mathrm{Linear}(512, 128)$   & $65{,}664$      & $65{,}664$ \\
$\mathrm{Linear}(128, 2)$     & $258$           & $258$ \\
\midrule
Total                         & $2{,}434{,}946$ & $2{,}433{,}922$ \\
\bottomrule
\end{tabular}
\end{table}

\subsection{Transitions and Safeguards}
\label{sec:supp-safeguards}

\paragraph{Acceptance thresholds.}
Placement is accepted at $\kappa_c = 0.5$ for recoverable planar placement
and $\kappa_c = 0.05$ for irreversible container placement. A false
rejection on the recoverable side costs a re-grasp; on the irreversible side
it inserts an additional object and invalidates the count.

\paragraph{Rollback.}
A rejected recoverable placement returns the pointer to the grasp subgoal of
the same repetition, so that the grasp and placement are repeated as a unit.
A rejected grasp leaves the pointer unchanged. Irreversible placement admits
no rollback, which is why its acceptance threshold is permissive rather than
strict.

\paragraph{Deadlock bounds.}
Consecutive rejections are bounded at two for recoverable placement and
three for irreversible placement and for grasping. On reaching the bound the
pointer advances regardless of the verdict, so a persistently rejecting
verifier cannot stall an episode. A grasp whose load persists without a
qualifying lift for 80 control steps is released from the pending state by a
stuck timeout.

\paragraph{Pre-termination check.}
Before the terminal subgoal, and only when the preceding subgoal was a
recoverable placement, AGM re-verifies that placement with a visual
occupancy test that does not involve the learned head. The test intersects a
hue mask for the manipulated object with a dilated mask of the target region
and accepts at an overlap of twelve pixels. A failed check returns the
pointer to the grasp subgoal of that repetition.

\subsection{Hyperparameters}
\label{sec:supp-hparams}

Table~\ref{tab:hyperparams} collects every constant introduced above. The
set is small, and one configuration serves all four tasks and every
repetition count; nothing is selected per task, per count, or per episode.
Only one pair is deliberately asymmetric, the two acceptance thresholds
$\kappa_c$, which differ by an order of magnitude according to whether the
placement they judge can be undone. Table~\ref{tab:ablation} ablates
exactly that pair.

\begin{table}[htbp]
\centering
\caption{Complete simulation hyperparameter set, shared across all four
tasks and all episodes.}
\label{tab:hyperparams}
\small
\begin{tabular}{@{}llr@{}}
\toprule
Symbol & Quantity & Value \\
\midrule
\multicolumn{3}{@{}l}{\emph{Event localization}}\\
--- & Closure entry width & $0.030$ \\
--- & Closure exit width & $0.035$ \\
--- & Empty-closure width & $0.008$ \\
--- & Hold-confirm & 5 frames \\
\midrule
\multicolumn{3}{@{}l}{\emph{Grasp verification}}\\
$N_q$ & Query grid & $6 \times 6$ \\
--- & Query half-width & $22$\,px \\
$\eta_{\mathrm{vis}}$ & Visibility threshold & $0.5$ \\
$\rho$ & Retained fraction & $1/3$ (min.\ 6 pts) \\
$\delta_G$ & Coherent-lift threshold & $4$\,px \\
--- & Lift closing the clip & $0.03$\,m \\
--- & Stuck timeout & 80 steps \\
\midrule
\multicolumn{3}{@{}l}{\emph{Placement verification}}\\
$K$ & Frames per window & 4 \\
--- & Settling, recoverable & 30 frames \\
--- & Settling, irreversible & 40 frames \\
$\kappa_c$ & Acceptance, recoverable & $0.5$ \\
$\kappa_c$ & Acceptance, irreversible & $0.05$ \\
--- & Feature width & $4626$ \\
\midrule
\multicolumn{3}{@{}l}{\emph{Transitions}}\\
--- & Rejection bound, recoverable & 2 \\
--- & Rejection bound, irreversible & 3 \\
--- & Rejection bound, grasp & 3 \\
--- & Occupancy overlap & 12\,px \\
\bottomrule
\end{tabular}
\end{table}

\section{Verifier Training and Analysis}
\label{sec:supp-verifier}

Every write into memory rests on a verification decision, so the head that
makes those decisions deserves its own account: where its training events
come from, how it is fitted, and what it can and cannot discriminate.

\subsection{Event Dataset}
\label{sec:supp-events}

The verification head is trained on 788 execution events collected on the
official RoboMME training split. Table~\ref{tab:events-type} gives the
composition by achievement type, Table~\ref{tab:events-ctrl} by collection
controller.

\begin{table}[htbp]
\centering
\caption{The 788 training events by achievement type. Failures constitute
25.1\% of the corpus.}
\label{tab:events-type}
\small
\begin{tabular}{@{}lrrr@{}}
\toprule
Type & Total & Achieved & Failed \\
\midrule
\textsf{grasp}       & 384 & 282 & 102 \\
\textsf{place-rev}   & 278 & 221 & \phantom{0}57 \\
\textsf{place-irrev} & 126 & \phantom{0}87 & \phantom{0}39 \\
\midrule
Total & 788 & 590 & 198 \\
\bottomrule
\end{tabular}
\end{table}

\begin{table}[htbp]
\centering
\caption{The same events by collection controller, over 61, 40, and 40
episodes respectively; both PickXTimes controllers draw on the same
100-seed training split, so their episode sets overlap.}
\label{tab:events-ctrl}
\small
\begin{tabular}{@{}llrr@{}}
\toprule
Controller & Type & Ach. & Fail \\
\midrule
AGM (PickXTimes)      & \textsf{grasp}       & 137 & \phantom{0}27 \\
                      & \textsf{place-rev}   & 165 & \phantom{00}0 \\
Attempt-based (PickX) & \textsf{grasp}       & \phantom{0}40 & \phantom{0}72 \\
                      & \textsf{place-rev}   & \phantom{0}56 & \phantom{0}57 \\
AGM (BinFill)         & \textsf{grasp}       & 105 & \phantom{00}3 \\
                      & \textsf{place-irrev} & \phantom{0}87 & \phantom{0}39 \\
\bottomrule
\end{tabular}
\end{table}

Table~\ref{tab:events-ctrl} explains why two controllers are needed. Under
AGM, planar placement on PickXTimes yields 165 successes and no failures at
all: a controller that verifies its own progress rarely leaves failures in
its trace, so training on its data alone would present the head with an
almost purely positive corpus. The attempt-based controller, which advances
without verification, contributes 72 failed grasps and 57 failed placements
and supplies most of the negative evidence the head learns from.

\paragraph{Seed disjointness.}
Collection draws on the benchmark \textsf{train} split: 100 PickXTimes
episodes with seeds in $[1000, 10900]$ and 100 BinFill episodes with seeds
in $[4000, 13902]$. All reported evaluation draws on the \textsf{test}
split, 50 episodes per task with seeds in $[510000, 514900]$ and
$[540000, 544900]$. We verified programmatically that the two seed sets are
disjoint for both tasks, so no evaluation episode contributes to verifier
training.

\paragraph{Label predicates.}
Environment predicates construct offline labels and are unavailable at
evaluation. A grasp is labeled achieved when the object rises above
$0.05$\,m with an active gripper--object contact predicate; a recoverable
placement when the object lies within $0.05$\,m of the target center and has
left the gripper; an irreversible placement when the container occupancy
count increases. Post-release windows are sampled at the deployment
verification time, 40 frames for container placement and 30 otherwise.

\subsection{Training Configuration}
\label{sec:supp-training}

The head is
$\mathrm{Linear}(4626, 512)$--ReLU--$\mathrm{Dropout}(0.3)$--$\mathrm{Linear}(512, 128)$--ReLU--$\mathrm{Linear}(128, 2)$,
optimized with AdamW at learning rate $10^{-3}$ and weight decay $10^{-4}$
for 300 full-batch gradient steps. Class-weighted cross-entropy compensates
for label imbalance, with weights set to inverse class frequency and
normalized to sum to the number of classes. No early stopping is applied.

Grasp and placement events share the feature map of Eq.~\eqref{eq:event_feature} and train
jointly in one head. At deployment the grasp branch never queries that head,
since grasp achievement is decided by point tracking, so grasp events act
purely as auxiliary supervision for the shared representation.

\paragraph{Measurement model and deployed model.}
Two models must be distinguished. For measurement, the events are
partitioned at the episode level under a fixed permutation of episode keys,
one fifth held out, giving 650 training and 138 held-out events; a head
fitted to the 650 is scored on the 138. For deployment, a head is then
fitted to all 788 events under identical settings, and this refitted head
produces every AGM result in the paper. All 788 events originate from the
training split and evaluation uses disjoint test-split seeds, so the refit
leaves the reported task success rates uncontaminated. It does mean the
held-out figures below characterize the training procedure rather than the
specific deployed weights.

\subsection{Held-Out Verification Performance}
\label{sec:supp-heldout}

Table~\ref{tab:verifier} reports the measurement model on the 138 held-out
events by achievement type, under a fixed seed. Metrics are given for the
\textsf{failed} class, since a missed failure is what writes false progress
into memory.

\begin{table}[htbp]
\centering
\caption{Held-out performance of the measurement model at $\mathrm{argmax}$,
in percent. Grasp events are included because they supply auxiliary
supervision, though deployment decides grasps by point tracking rather than
by this head.}
\label{tab:verifier}
\small
\begin{tabular}{@{}lrrrrr@{}}
\toprule
Subset & $n$ & Acc. & \multicolumn{3}{c}{\textsf{failed} class} \\
\cmidrule(l){4-6}
 & & & P & R & F1 \\
\midrule
All                  & 138 & 91.3 & 87.5 & 70.0 & 77.8 \\
\textsf{grasp}       & \phantom{0}69 & 92.8 & 100.0 & 68.8 & 81.5 \\
\textsf{place-rev}   & \phantom{0}41 & 100.0 & 100.0 & 100.0 & 100.0 \\
\textsf{place-irrev} & \phantom{0}28 & 75.0 & \multicolumn{3}{c}{see text} \\
\bottomrule
\end{tabular}
\end{table}

The two placement types behave very differently, and that difference is what
the threshold split of Appendix~\ref{sec:supp-safeguards} responds to.

On recoverable planar placement the head is exact. All 41 held-out events
are classified correctly at the deployment threshold of $0.5$, failures
included. This is the branch whose rejections are acted on strictly, and on
which a rejection returns the pointer to the preceding grasp.

On irreversible container placement the head is not a reliable
discriminator. The held-out split holds 28 such events of which four are
failures, and at $\mathrm{argmax}$ none of the four is caught. Four
negatives are too few to estimate discrimination with confidence, but the
direction is unambiguous, and it matches how the branch is deployed:
acceptance at $\kappa_c = 0.05$ does not ask the head to separate successes
from failures, only to withhold acceptance when it is confidently negative.
That happened five times in one of the two reported BinFill runs, at scores between
$0.000$ and $0.044$.

The design does not rest on discrimination here, and Table~\ref{tab:ablation} of the main
paper measures the cost of assuming that it should: flattening the split so
that container placement is judged strictly drops BinFill from 84.0\% to
74.0\%, because strict acceptance rejects correct drops and stalls the
pointer on a subgoal already complete. Verification earns its accuracy on
the recoverable side; on the irreversible side its role is to avoid
destroying a count that is already correct.

\subsection{Alignment of the Post-Release Window}
\label{sec:supp-window}

The post-release window is sampled at exactly the deployment verification
time. An earlier configuration in which the collection offset exceeded the
deployment offset by 20 frames reached 66.0\% on BinFill, against 84.0\%
(mean of two runs) under matched offsets. The two configurations also
differed in the verifier's training corpus, which was larger and drawn in
part from the longer window, so we present the gap as the observation that
motivated aligning the offsets rather than as an isolated measurement of the
offset alone. The mechanism is nonetheless specific: a container placement
photographed too late shows a settled scene whose appearance no longer
distinguishes a successful drop from an object that entered and rebounded,
and the head loses the very evidence it was trained to read.

\section{Simulation Evaluation}
\label{sec:supp-sim}

Below are the protocol behind Table~\ref{tab:main} and Table~\ref{tab:ablation}, a controlled comparison
isolating what verified writing contributes, and complete rollouts at the
deepest repetition count in the benchmark.

\subsection{Protocol}
\label{sec:supp-protocol}

Evaluation uses the official 50-episode test split for each of the four
RoboMME Counting tasks, with the seed ranges of Appendix~\ref{sec:supp-events}. An episode
succeeds only if the robot presses the stop button after exactly the
instructed count within the 1{,}300-step budget; a correct count followed by
no termination, and a termination after the wrong count, both score as
failures. AGM entries average two independent runs that differ only in
random seed, with identical thresholds, transition rules, and verifier
weights.

Verification runs out of process. The point tracker and the verification
head are hosted as separate services and queried over a local socket, so the
evaluation process itself loads neither the tracker nor the vision-language
encoder.

\paragraph{Substituting the instruction-only plan.}
The subgoal sequence is read from the benchmark decomposition by default. To
check that AGM does not depend on privileged structure, we also run the full
test split with plans produced by a parser that reads only the instruction
string and touches neither environment state nor task ground truth. It
yields 98.0\% on PickXTimes and 82.0\% on BinFill, against 100.0\% and
84.0\% with the benchmark decomposition. The parser emits canonical subgoal
wording that differs verbatim from the benchmark phrasing while preserving
the type, object, and length of the sequence, so the comparison bounds how
much of AGM's performance is attributable to the provided decomposition
rather than to the memory discipline.

\subsection{What Verified Writing Contributes}
\label{sec:supp-source}

The main text argues that a progress memory advancing on every attempt can
be worse than carrying no progress memory at all. Table~\ref{tab:source}
isolates that comparison. Three configurations share one subgoal interface
and differ only in what decides the current subgoal: AGM, which advances on
verified achievement; an attempt-based controller, identical to AGM except
that both verification branches are switched off, so that any gripper
open--close cycle advances the pointer; and a control that receives the whole
instruction for the entire episode and has no pointer.

\begin{table}[htbp]
\centering
\caption{Success rate (\%) by what decides the current subgoal, on the
official 50-episode test split. AGM and the attempt-based controller share
one subgoal-conditioned checkpoint and differ only in whether the pointer
waits for evidence. $\dagger$: a policy trained for
full-instruction conditioning rather than the released checkpoint of
Table~\ref{tab:main}, since a subgoal-conditioned checkpoint given an entire instruction
is out of distribution and cannot serve as this control. AGM entries average
two runs; the other two are single runs.}
\label{tab:source}
\small
\begin{tabular}{@{}lrr@{}}
\toprule
Subgoal source & PickXTimes & BinFill \\
\midrule
AGM (verified writes) & \textbf{100.0} & \textbf{84.0} \\
Attempt-based (unverified writes) & 32.0 & 82.0 \\
Whole instruction (no pointer)$^\dagger$ & 36.0 & 40.0 \\
\bottomrule
\end{tabular}
\end{table}

On PickXTimes the gap is decisive. The same policy, the same subgoal
sequence, and the same pointer reach 100.0\% when writes wait for evidence
and 32.0\% when they do not, and unverified writing does not improve on
handing the policy the whole instruction. Decomposition is not what produces
the result. A pointer that advances on attempts converts every transient
grasp or placement failure into a permanent counting error, and on a task
where the object stays retrievable those errors accumulate faster than the
decomposition helps.

On BinFill the attempt-based controller nearly matches AGM. The explanation
is structural: a cube released into the container is removed from the scene,
so an attempt that reaches release is almost always an achievement, and the
two quantities that AGM distinguishes very nearly coincide. There is little
for verification to separate, which is the same fact that Appendix~\ref{sec:supp-heldout}
measures from the other direction, where the head has no reliable
discrimination on container placement. What the case split still buys on
this task is protection against false rejection, worth ten points by
Table~\ref{tab:ablation}.

The two tasks together locate the mechanism rather than merely confirming
it. Verified writing pays in proportion to how often an attempt fails to
become an achievement. Where a failed action leaves the world recoverable,
that gap is wide and verification decides the outcome; where the action is
irreversible and self-evidencing, the gap narrows and the decomposition
carries most of the result.

\subsection{Qualitative Rollouts}
\label{sec:supp-sim-qual}

\begin{figure}[t]
\centering
\includegraphics[width=\textwidth]{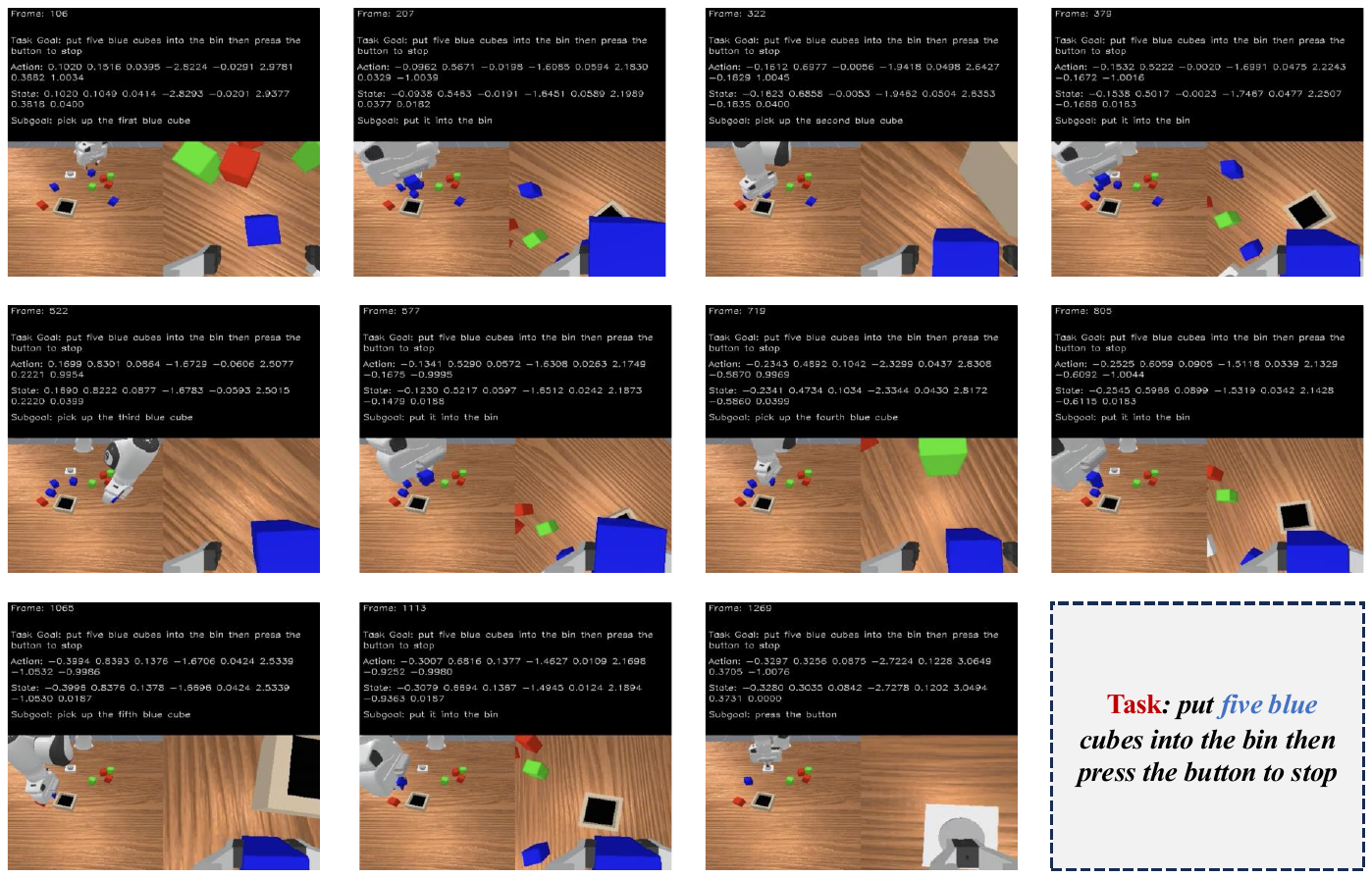}
\caption{Simulation rollout of BinFill at $N{=}5$ under AGM. One panel per
pointer transition, overlaid with the frame index, task instruction, action
and state vectors, and the active subgoal. The eleven subgoals---five
grasp--placement pairs and the terminal button press---are issued in order,
and the pointer advances only on a verified achievement, so the episode
terminates after exactly five cubes have entered the bin.}
\label{fig:sim-binfill}
\end{figure}

\begin{figure}[t]
\centering
\includegraphics[width=\textwidth]{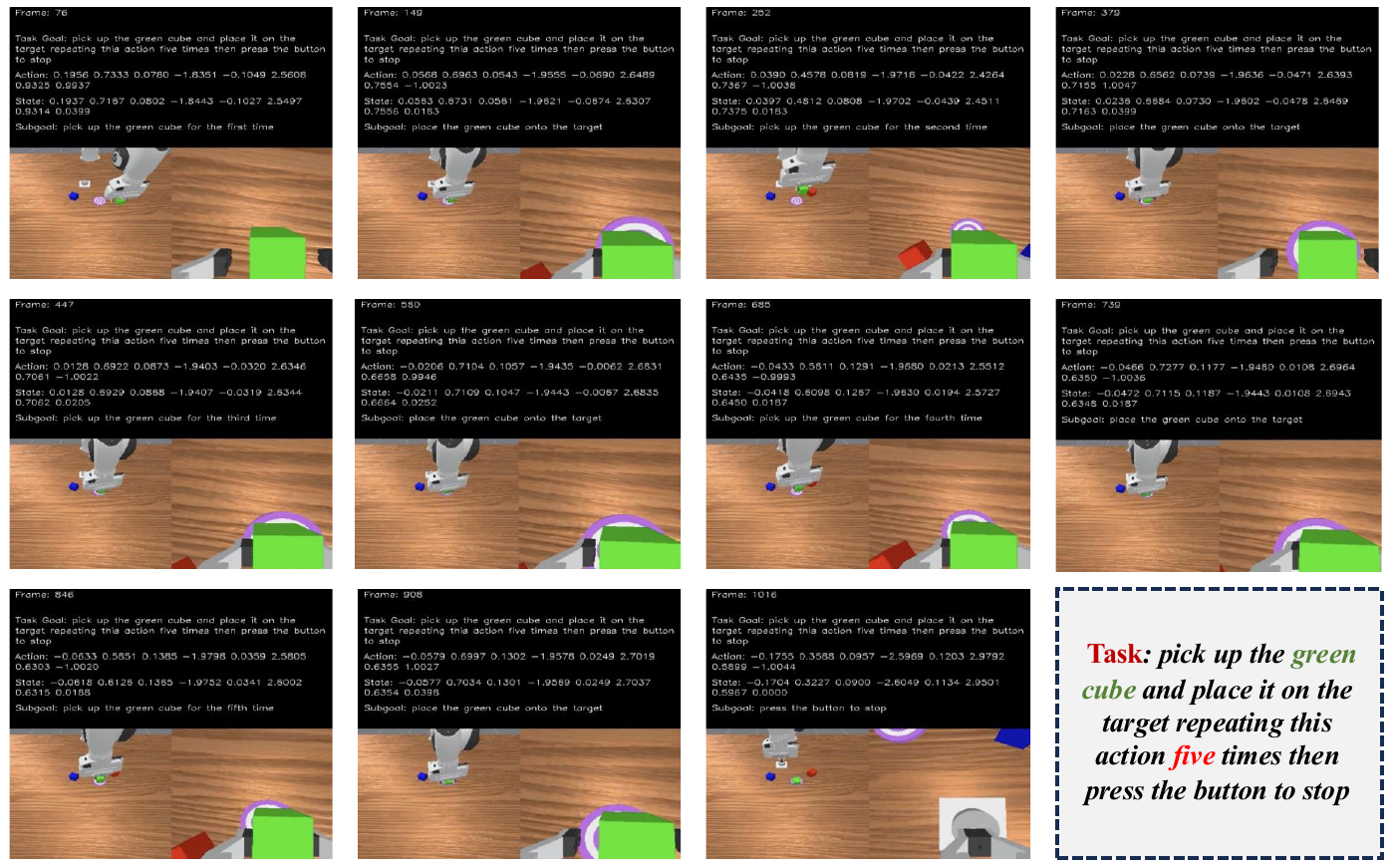}
\caption{Simulation rollout of PickXTimes at $N{=}5$ under AGM, green cube.
The same cube is grasped and placed five times, so the scene at every grasp
is visually near-identical; the repetition index appears only in the subgoal
supplied by the pointer.}
\label{fig:sim-pickx-green}
\end{figure}

\begin{figure}[t]
\centering
\includegraphics[width=\textwidth]{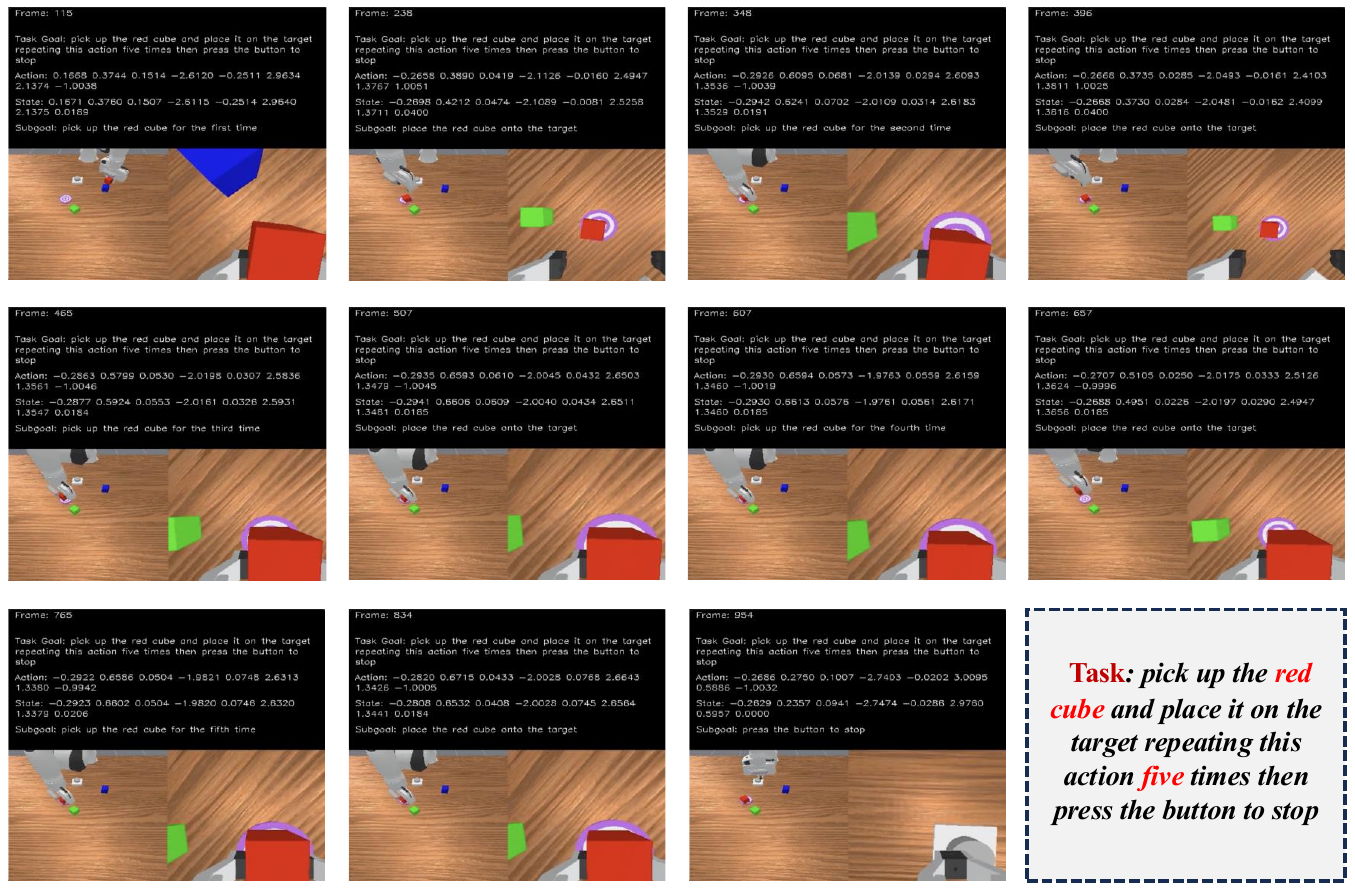}
\caption{Simulation rollout of PickXTimes at $N{=}5$ under AGM, red cube.
The instruction, object, and target differ from
Figure~\ref{fig:sim-pickx-green}, while thresholds, transition rules, and
verifier weights are identical.}
\label{fig:sim-pickx-red}
\end{figure}

Figures~\ref{fig:sim-binfill}--\ref{fig:sim-pickx-red} show complete
simulation episodes at $N{=}5$, the deepest repetition count in the
benchmark and the one on which the frozen policy scores zero. Each panel is
captured at a pointer transition and carries the frame index, the task
instruction, the current action and state vectors, and the subgoal in force
at that instant.

The two PickXTimes rollouts make the aliasing problem visible. Panels 1, 3,
5, 7 and 9 of Figure~\ref{fig:sim-pickx-green} are the same scene: one cube
on the table, one empty target, one gripper approaching. Nothing in any of
those images indicates which repetition is underway, and a policy that must
recover that fact from its current observation has nothing to recover it
from. The subgoal printed in each overlay is supplied by the pointer, and
the pointer is what distinguishes the five otherwise indistinguishable
states.

\section{Real-Robot Deployment}
\label{sec:supp-real}

This section documents the physical platform, demonstration collection,
automatic subgoal annotation, policy training, the adaptation of the AGM
runtime to hardware, and the evaluation protocol behind the \emph{Real}
block of Table~\ref{tab:perN}.

The AGM mechanism transfers unchanged. The task memory
$\mathcal{M}_t = (\mathcal{G}, p_t)$, the state machine of Eq.~\eqref{eq:fsm}, the
verification of Eqs.~\eqref{eq:grasp_motion_score} and~\eqref{eq:placement_score}, the transition rule of Eq.~\eqref{eq:memory_transition}, the
deadlock bounds, and the pre-termination check are ported line for line.
Modifications are confined to the interface layer: sensor dimensionality,
camera geometry, threshold recalibration, and service reliability. Each is
stated below.

\subsection{Physical Platform}
\label{sec:supp-platform}

\paragraph{Robot.}
The follower arm is a 6-DoF AgileX PiPER manipulator with a 1-DoF parallel
gripper of commanded stroke $0$--$0.08$\,m, controlled at 30\,Hz over a CAN
bus. State and action spaces are 7-dimensional: six joint positions in
radians plus gripper width in meters. The \emph{measured} gripper width, as
opposed to the commanded width, is exposed as an independent channel and
supplies $w_t$ to the state machine.

\paragraph{Cameras.}
Two Intel RealSense D435 RGB-D cameras run at $640 \times 480$ and 30\,FPS: a
front camera mounted above the workspace (third-person, eye-to-hand) and a
wrist camera on the end-effector (eye-in-hand). The policy consumes both RGB
streams; aligned depth is recorded but unused. The front intrinsics $K$ and
eye-to-hand extrinsics $E$ of Eq.~\eqref{eq:grasp_projection} are calibrated with a ChArUco target,
with residuals near $23$\,mm in translation and $5.4^\circ$ in rotation, a
reprojection error of roughly $15$--$40$\,px. That suffices to seed the query
grid of Section~\ref{sec:grasp_verification}, and nothing downstream requires more: calibration decides
where to look, while the achievement decision reads relative point motion
inside the tracked window.

\paragraph{Distributed inference.}
Inference runs on a GPU workstation separate from the robot host, connected
by a WebSocket link tunneled over SSH. The robot host runs a thin client
handling camera capture, the arm driver, and the safety layer. Each control
frame, the client sends one observation message (JPEG at quality 90, about
3\,MB/s at 30\,Hz) to the orchestration service, which advances the pointer,
rebuilds the language conditioning for the current subgoal, requests a new
action chunk when the client's chunk is exhausted or the subgoal has
changed, and returns the subgoal together with any new actions and the stop
flag. The placement head and the point-tracking channel run as two further
services, called synchronously. When the subgoal switches, the client
discards its remaining actions and executes the chunk generated under the
new conditioning, reproducing the actuator re-prime of the simulation
implementation.

\subsection{Task Suite}
\label{sec:supp-tasks}

Two long-horizon families from the benchmark are reproduced on hardware
using 26\,mm colored plastic cubes, a bin, a planar target mat, and a
tabletop button whose depression ends the episode:

\begin{itemize}
\item \textbf{BinFill}$(N, \text{colors})$ --- \emph{``put $N$ \{color\}
      cubes into the bin, then press the button to stop.''}
\item \textbf{PickXTimes}$(N, \text{color})$ --- \emph{``pick up the
      \{color\} cube and place it on the target, repeating this action $N$
      times, then press the button to stop.''} The same cube is grasped and
      placed $N$ times, with the operator returning it to the workspace
      between repetitions per the benchmark's reset convention.
\end{itemize}

Each task expands into
$\mathcal{G} = [\textsf{pick}, \textsf{place}] \times N + [\textsf{button}]$,
that is $T = 2N{+}1$ subgoals, up to eleven at $N{=}5$. Subgoal wording is
taken verbatim from the benchmark templates, which matters for
train--deployment consistency (Appendix~\ref{sec:supp-policy}).

\paragraph{Demonstrated and evaluated repetition depth.}
Single-color BinFill demonstrations stop at $N{=}4$, PickXTimes
demonstrations reach $N{=}5$, and evaluation spans $N = 1, \dots, 5$ for
both. Of the BinFill $N{=}5$ trials, $60\%$ instruct a single color and
therefore evaluate a configuration absent from the demonstration corpus;
Appendix~\ref{sec:supp-depth} analyzes what follows. The task and subgoal vocabularies are enumerated programmatically
from the templates over $N = 1, \dots, 5$ rather than harvested from the
corpus, so every string emitted at $N{=}5$ remains in-vocabulary for the
parity check of Appendix~\ref{sec:supp-policy}, and the generalization at issue is visuomotor
rather than lexical.

\subsection{Demonstrations and Automatic Subgoal Annotation}
\label{sec:supp-demos}

\paragraph{Teleoperation.}
Demonstrations are collected by leader--follower teleoperation, the leader a
second kinematically identical PiPER. Joint targets pass through a one-euro
filter with minimum cutoff $0.4$\,Hz and $\beta = 5.0$. The gripper channel
is left unfiltered to preserve grasp timing, and the leader gripper range
maps linearly onto the follower's $0$--$0.08$\,m stroke.

\paragraph{Corpus.}
194 episodes, 100 BinFill and 94 PickXTimes, totaling 237{,}517 frames or
roughly 2.2 hours at 30\,Hz, each with both camera streams, 7-dimensional
proprioception, and end-effector pose.

\paragraph{Segmentation.}
Episodes are segmented programmatically from the gripper signal rather than
by hand or by a vision-language model, mirroring the event logic AGM uses at
deployment. Commanded gripper edges mark operator intent; a measured width
plateau consistent with a held cube and lasting five frames confirms a
grasp; release events close each cycle. A plateau-width check at
$0.040$\,m rejects closures on empty air. The expected subgoal sequence is
parsed from the task string and the number of detected cycles reconciled
against $N$, which closes the loop as a quality gate. Color attribution in
multi-color BinFill episodes uses HSV statistics from the wrist camera at
the grasp instant.

Boundaries are placed at the confirmation frame, the instant at which the
deployed verifier would advance the pointer, making the annotation the
offline analogue of AGM's online transition.

\paragraph{Audit.}
An independent visual review pipeline covered all 194 episodes, sampling
2{,}442 frames alongside programmatic and signal-level re-derivation. All
194 episodes were correctly labeled, with pick boundaries within 4 frames
(PickXTimes) and 7 frames (BinFill) of the independent visual estimate at
30\,Hz.

\subsection{Policy Training and Language Conditioning}
\label{sec:supp-policy}

Two architectures, $\pi_0$ and $\pi_{0.5}$, are fine-tuned under two
conditioning formats, giving the four checkpoints of Table~\ref{tab:perN}. The
\emph{fused} checkpoints train on the frame-level prompt below, whose text
channel carries the full task and the active subgoal together; these are the
AGM-compatible policies. The \emph{merged} checkpoints train on the same
episodes with each frame's prompt reduced to the bare task instruction, and
appear in Table~\ref{tab:perN} as $\pi_0$ and $\pi_{0.5}$. All four are full fine-tunes
from the official base checkpoints with the vision encoder frozen, under
identical optimization settings and the same 194 episodes, so that
$\pi_0$ against $\pi_{0.5}$ and fused against merged are both controlled
comparisons. The training budget matches that of the benchmark's released
80{,}000-step checkpoint: we train for 40{,}000 steps at twice its batch
size.

\begin{table}[htbp]
\centering
\caption{Policy fine-tuning configuration for the real-robot experiments.
$\pi_{0.5}$ uses quantile normalization, $\pi_0$ uses $z$-score
normalization, following each architecture's default.}
\label{tab:real-train}
\small
\begin{tabular}{@{}lp{0.50\columnwidth}@{}}
\toprule
Hyperparameter & Value (identical across all four) \\
\midrule
Training steps   & 40{,}000 \\
Batch size       & 128 \\
Optimizer        & AdamW, gradient-norm clip $1.0$ \\
Learning rate    & linear warmup to $5\times10^{-5}$ over 10k steps,
                   constant thereafter \\
EMA decay        & $0.999$ \\
Action horizon   & 20 frames \\
Vision encoder   & frozen (SigLIP) \\
Normalization    & architecture default, statistics computed on this
                   corpus \\
\bottomrule
\end{tabular}
\end{table}

Observations are the front and wrist images letterboxed to
$224 \times 224$ together with the 7-dimensional proprioceptive state;
actions are 7-dimensional joint-plus-gripper targets emitted in chunks of 20
frames, of which the client executes 10 ($0.33$\,s) before replanning.

\paragraph{Fused prompt.}
Frame-level annotations enter training as

\begin{quote}
\ttfamily\small
Task: \{full task instruction\}\\
Current Subgoal: \{current subgoal\}.
\end{quote}

\noindent
so that one text channel carries global context and the active subgoal. At
deployment the orchestrator rebuilds this string from $g_{p_t}$ through the
same template module used for annotation, a single source of truth for all
wording. A parity check before every run requires that every string the
deployment can emit appear verbatim in the training vocabulary, which makes
the conditioning distribution at deployment provably identical to the one
seen in training. We recommend the check as general practice for
subgoal-conditioned deployment, since conditioning drift is otherwise
silent.

\subsection{AGM Runtime on Hardware}
\label{sec:supp-runtime}

Only the placement head is retrained. The remaining changes recalibrate
constants whose simulation values are expressed in units that do not carry
over to the physical platform.

\subsubsection{State Machine and Calibration}
\label{sec:supp-fsm-real}

The state machine transfers unchanged, five-frame hold-confirm included.
Only its thresholds are recalibrated to the physical gripper by a scripted
probe over three configurations: cube held, empty closure, fully open.

\begin{table}[htbp]
\centering
\caption{State machine calibration on the physical platform.}
\label{tab:real-fsm}
\small
\begin{tabular}{@{}lr@{}}
\toprule
Quantity & Calibrated value \\
\midrule
Measured width, fully open          & $0.0692$\,m \\
Measured width, holding 26\,mm cube & $0.0260$\,m \\
Measured width, empty closure       & $0.0031$\,m \\
Closure entry                       & $0.0476$\,m \\
Closure exit                        & $0.0541$\,m \\
Empty-closure threshold             & $0.0146$\,m \\
Hold-confirm                        & 5 frames \\
Plateau sanity width                & $0.040$\,m \\
\bottomrule
\end{tabular}
\end{table}

The same probe registers the bin, target, and button regions in the robot
base frame as axis-aligned rectangles, with a press height for the button,
used by the proprioceptive gate of Appendix~\ref{sec:supp-placement-real}.

\subsubsection{Grasp Verification}
\label{sec:supp-grasp-real}

A pick commits only once verified, and a release cancels any pending pick,
so an empty closure can never advance the pointer.

The proprioceptive channel requires the end-effector to rise $0.03$\,m while
the load persists. The point-tracking channel ports Eqs.~\eqref{eq:grasp_projection}--\eqref{eq:grasp_motion_score}
unmodified and rescales only their geometric quantities. The grasp-instant
end-effector position projects into the front camera through the calibrated
$K$ and $E$, with a $0.12$\,m flange-to-fingertip offset, and a
$6 \times 6$ grid is seeded within a half-width of $60$\,px. CoTracker3
tracks the clip from the grasp to the present frame, and the pick is
confirmed when $r^G_n$ exceeds $\delta_G = 8$\,px. The pair $(8, 60)$ rescales the simulation
values $(4, 22)$: focal lengths are $f_x \approx 610$ at $640\times480$ on hardware
against $128$ at $256^2$ in simulation, and a $0.03$\,m lift subtends about
$16$\,px at our working distance.

Repeated rejection is bounded as in simulation, and a stuck timeout at
$2.7$\,s of held load without lift provides the final fallback.

\subsubsection{Placement Verification}
\label{sec:supp-placement-real}

The placement head is the only learned component retrained for hardware,
because its feature layout depends on proprioceptive dimensionality. The
architecture is unchanged: a frozen SigLIP (base-patch16-224) encodes
mean-pooled windows of $K = 4$ frames before and after release from both
cameras, and the head consumes the feature map of Eq.~\eqref{eq:event_feature},

\begin{equation*}
\begin{aligned}
f_n = \bigl[\;
& z^-_{f,n};\; z^+_{f,n};\; z^+_{f,n} - z^-_{f,n};\\
& z^-_{w,n};\; z^+_{w,n};\; z^{\mathrm{prop}}_n;\; z_{q,n}
\;\bigr].
\end{aligned}
\end{equation*}

\noindent
With $d = 7$ on hardware, $z^{\mathrm{prop}}_n$ has width 16 and the input
width is $4624$. Training uses class-weighted cross-entropy under an
episode-level split, and the text embedding $z_{q,n}$ uses the placement
subgoal wording, identical across collection, training, and serving.

\paragraph{Training data.}
Positives are 526 placement events, 238 on the target and 288 in the bin,
extracted automatically from the 194 demonstration episodes by the same
state machine used at deployment. Negatives are 75 events teleoperated to
fail on purpose across 20 dedicated episodes, spanning near-misses at the
target edge, knock-offs of a correctly placed cube, rebounds off the bin
rim, and gross misses; roughly 84\% are hard negatives that resemble
successes visually. Every negative was checked by human review of the
seconds around release, with accidental successes relabeled. The training
mixture keeps all 75 negatives and subsamples 175 positives, a 30\% failure
rate.

\paragraph{Decision rule.}
Target placement requires $s^P_n \ge 0.5$. Bin placement uses the permissive
threshold $0.05$ together with a proprioceptive gate requiring release with
the end-effector inside the registered bin region, since occlusion by the
bin makes the visual evidence unreliable there. This is the
recoverability-aware split of Appendix~\ref{sec:supp-safeguards} carried onto hardware. A rejected
placement rolls the pointer back to the associated grasp subgoal, bounded
consecutive rejections of two at the target and three at the bin force an
advance to break deadlock, and the pre-termination check re-verifies the
last placement before the terminal subgoal.

\subsubsection{Fail-Safe Semantics}
\label{sec:supp-failsafe}

Verification runs out of process in both settings; on hardware it also spans
two machines over a tunneled link, which widens the set of transport and
service faults that can interrupt a query. The physical deployment is
therefore fail-safe: a service fault returns a sentinel rather than a
verdict, whereupon AGM holds conservatively and retries after a cooldown,
with any genuine deadlock broken by the stuck timeout. A verdict is never
synthesized from a failed query. Services are probed at episode start, and
episodes in which a fault occurs are voided rather than continued, so no
reported episode rests on a degraded verification path.

\subsubsection{Safety Layer}
\label{sec:supp-safety}

All arms share one safety layer: a $1.5$\,s ramp from the current
configuration to the first policy action at episode start, activated when
that action deviates by $0.05$\,rad or more; a per-frame joint increment
limit of $0.1$\,rad; and hard assertions on action and state dimensionality.
The button subgoal waits at most $20$\,s, after which the episode terminates
and the subgoal scores as incomplete.

\subsection{Offline Replay Validation}
\label{sec:supp-replay}

Before any hardware rollout, all 194 demonstration episodes were replayed
through the deployment controller, with real gripper signals, real
end-effector trajectories, and real timing, and its behavior compared
against the ground-truth annotation.

\begin{table}[htbp]
\centering
\caption{Offline replay of the deployment controller on all 194
demonstration episodes, consuming no robot time.}
\label{tab:replay}
\small
\begin{tabular}{@{}lr@{}}
\toprule
Check & Result \\
\midrule
$\mathcal{G}$ matches annotated sequence   & 194 / 194 \\
Pointer reaches terminal subgoal           & 194 / 194 \\
Episode \textsf{stop} emitted              & 192 / 194 \\
Median transition offset                   & 7.5 frames (0.25\,s) \\
Transitions within 15 frames (0.5\,s)      & 96.5\% \\
\bottomrule
\end{tabular}
\end{table}

Achievement-grounded pointer advancement thus operates correctly on real
sensor signals before any hardware time is spent, and we recommend the
procedure as a bring-up practice for closed-loop embodied systems.

\subsection{Evaluation Protocol}
\label{sec:supp-eval}

Each backbone is evaluated in two configurations differing only in whether
the subgoal comes from AGM. The bare policy receives the full task
instruction for the whole episode; the AGM configuration receives the
subgoal $g_{p_t}$ indexed by the verified pointer. Both enter through an
identical conditioning interface under a matched data budget
(Appendix~\ref{sec:supp-policy}), so the difference is attributable to the source of the
subgoal.

Episodes are capped at 240\,s at 30\,Hz. The operator resets the scene
between episodes under a written protocol, re-randomizing cubes, target and
distractor positions within the workspace while the bin and button stay at
their calibrated poses. Initial layouts, cube counts, and color mixtures are
chosen to differ from the demonstration scenes, and $40\%$ of the BinFill
trials are dynamic: following the benchmark's dynamic-refresh rule, the
operator adds cubes to the workspace while the episode is running. Each episode logs a machine-readable trace,
covering every state-machine event, pointer transition, verifier query and
verdict, forced advance, and stop, all wall-clock stamped, alongside the
robot-view video and an independent third-person 4K recording used for
scoring. Success is scored per subgoal from the recording, and episode
success requires all $2N{+}1$ subgoals including the button press.

\paragraph{Mid-episode intervention.}
In the dynamic BinFill trials the operator adds cubes at moments that are
not restricted to the start of the episode: while the arm is closing on a
cube, while it is releasing a cube into the bin, and while it idles because
no cube of the instructed color is currently on the table. Each insertion
changes the table layout under the running policy and brings a human hand
into both camera views. Memory writes are considered only at the
gripper-load events emitted by the state machine from the measured gripper
width, so a hand entering the cameras' view or a cube appearing on the
table does not by itself open a verification. At those events the verifier
decides from the proprioceptive window, the pre- and post-event views of
both cameras, and the language query of the active subgoal, so it judges
the change in the instructed object--target relation rather than the
overall appearance of the scene. In the dynamic trials the interventions
caused no false acceptance, no false rejection, and no forced advance; the
frozen policy kept executing the current subgoal through them and proceeded
to grasp once a cube of the instructed color was available. The bottom row
of Figure~\ref{fig:real_rollout} shows such an episode.

\subsection{Repetition Depth Beyond the Demonstrations}
\label{sec:supp-depth}

Single-color BinFill at $N{=}5$ appears in no demonstration and makes up
$60\%$ of the BinFill $N{=}5$ trials, yet $\pi_{0.5}$+AGM reaches 60.0\% on
that cell against 70.0\% at the demonstrated $N{=}4$. The step across the
distribution boundary costs about as much as a step between adjacent
demonstrated counts, so crossing it adds no penalty beyond the physical
difficulty of a bin that is progressively more occupied.

This follows from externalizing progress into the pointer. The frozen policy
is conditioned on one subgoal at a time, and each subgoal of an $N{=}5$
episode exercises visuomotor skills seen in the demonstrations; only their
number is new,
and the number lives in memory rather than in the policy. Achievable
repetition depth is therefore governed by the memory rather than by the
counts present in the corpus, a property that follows from the interface and
is unavailable to any policy required to represent progress internally. The
bare $\pi_{0.5}$ makes the point from the other side: it scores zero at both
$N{=}4$ and $N{=}5$, failing well before the boundary is reached, so what
limits it is the missing task state rather than an unseen count.

PickXTimes is demonstrated through $N{=}5$, so its uniform 100.0\% is an
in-distribution result and the claim above is made for BinFill alone.

\subsection{Qualitative Results}
\label{sec:supp-real-qual}

\begin{figure}[t]
\centering
\includegraphics[width=\textwidth]{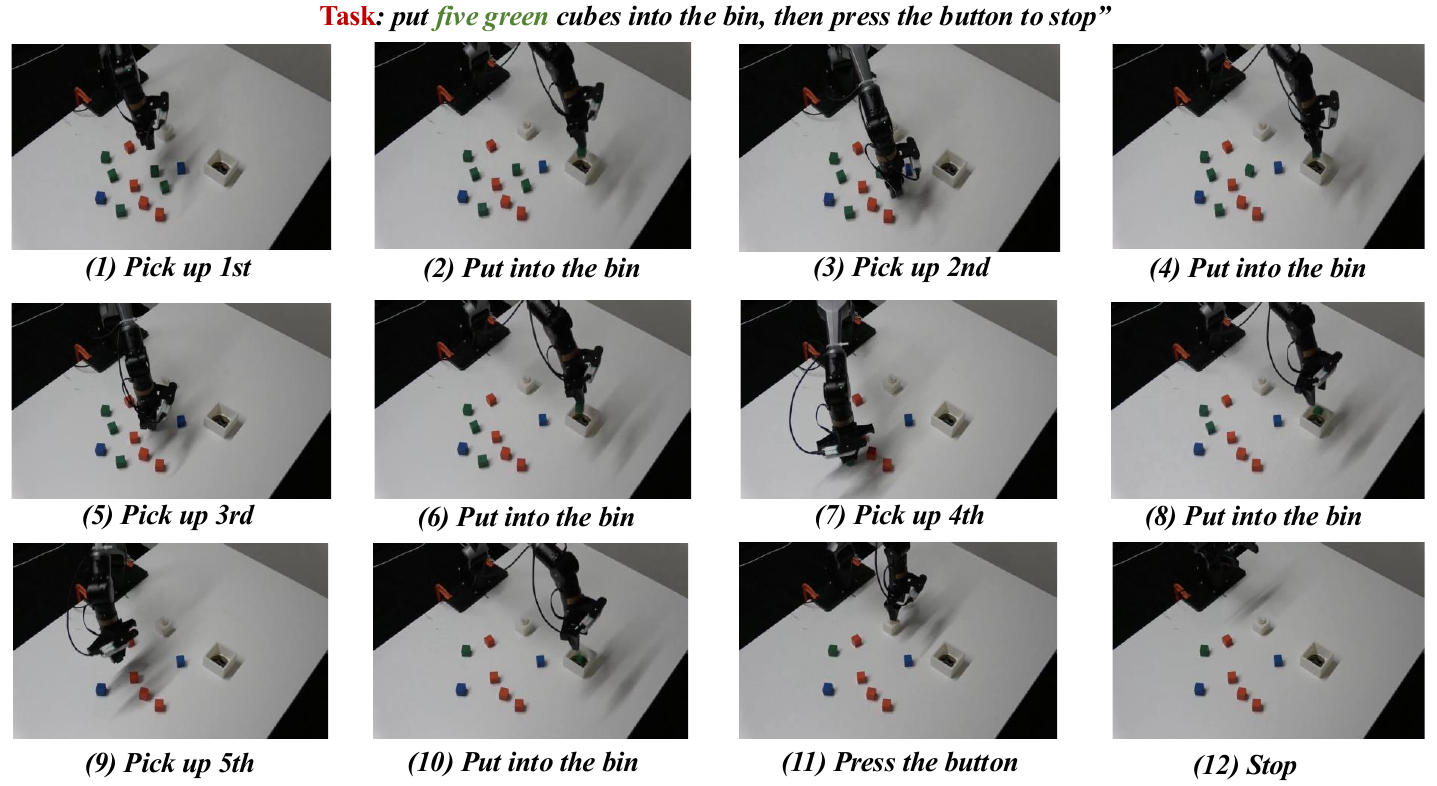}
\caption{Real-robot rollout of BinFill at $N{=}5$ with a single-color
instruction, a configuration absent from the demonstration corpus
(Appendix~\ref{sec:supp-tasks}). Cubes of three colors share the
workspace, so every grasp is color-selective while the pointer counts
deliveries of the instructed color. The episode terminates after exactly
five placements.}
\label{fig:real-binfill}
\end{figure}

\begin{figure}[t]
\centering
\includegraphics[width=\textwidth]{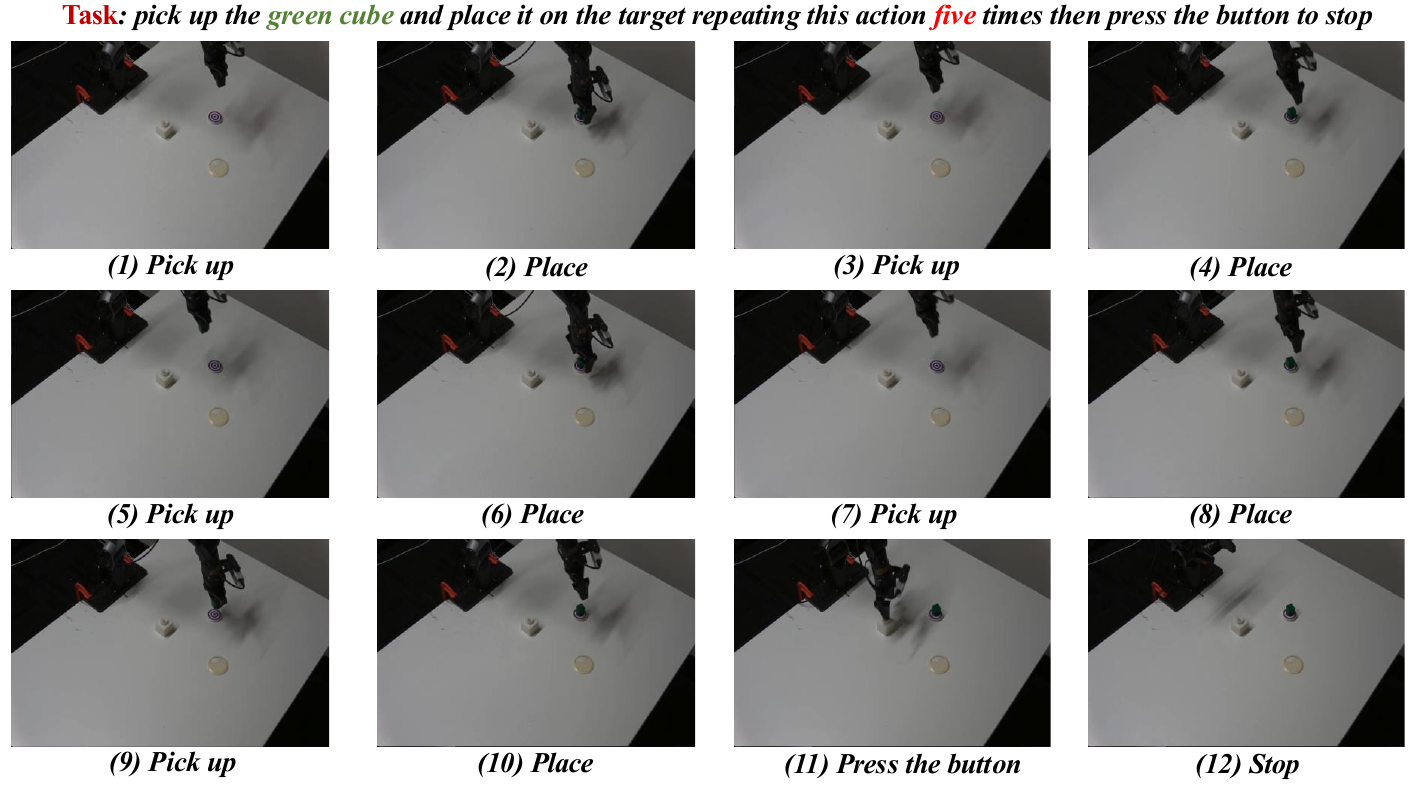}
\caption{Real-robot rollout of PickXTimes at $N{=}5$, green cube. The cube
returns to the workspace between repetitions, so panels 1, 3, 5, 7 and 9 are
the same scene; only the pointer distinguishes them.}
\label{fig:real-pickx-green}
\end{figure}

\begin{figure}[t]
\centering
\includegraphics[width=\textwidth]{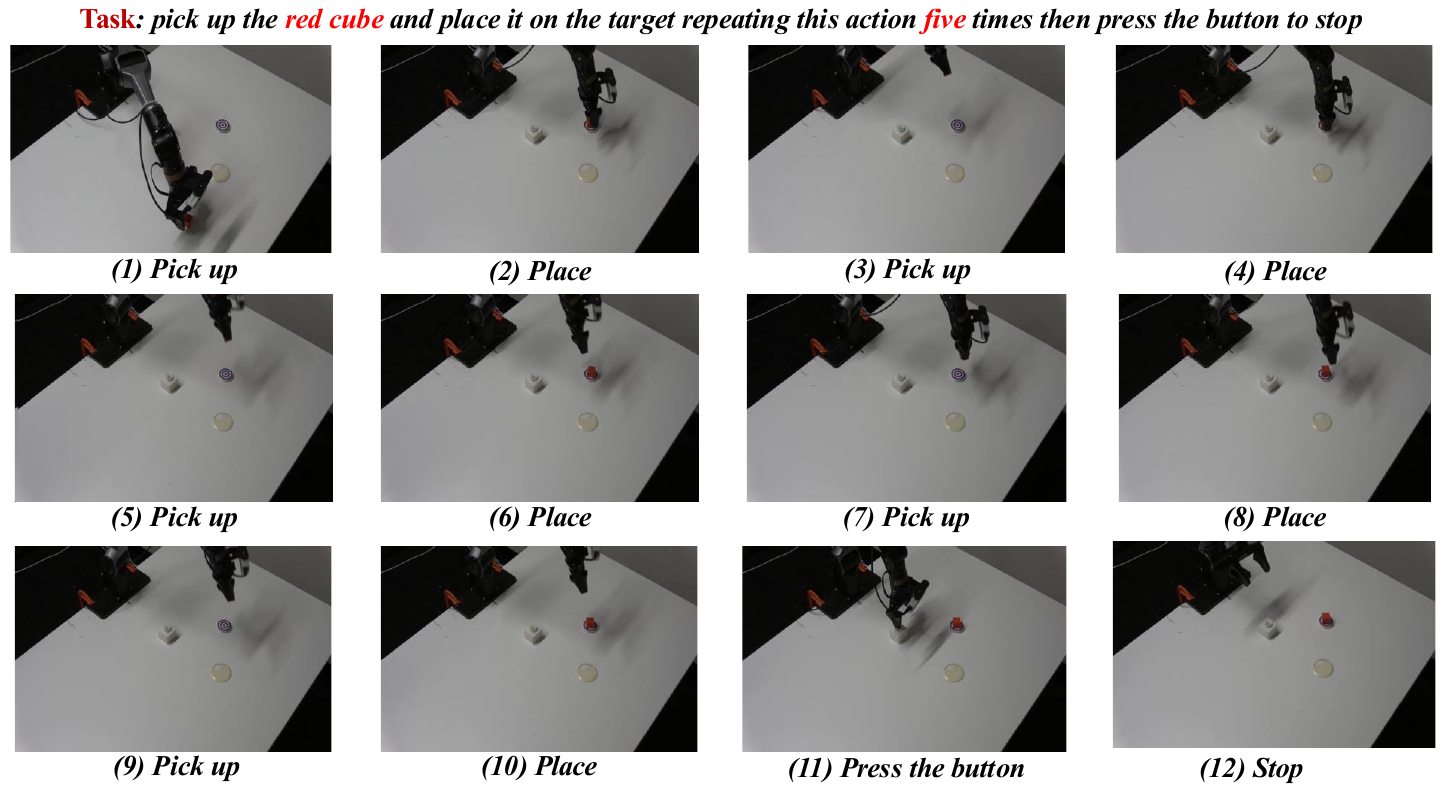}
\caption{Real-robot rollout of PickXTimes at $N{=}5$, red cube, under the
same verification head and thresholds as
Figure~\ref{fig:real-pickx-green}.}
\label{fig:real-pickx-red}
\end{figure}

Figures~\ref{fig:real-binfill}--\ref{fig:real-pickx-red} show complete
physical episodes at $N{=}5$, one panel per subgoal followed by the
terminated scene. All three succeed, and all three run the framework
configuration of Appendix~\ref{sec:supp-runtime} without modification.

Figure~\ref{fig:real-binfill} is the undemonstrated setting of
Appendix~\ref{sec:supp-depth}. The workspace holds cubes of three colors, so each grasp
selects by color while the pointer tracks how many of the instructed color
have been delivered; the two quantities are independent, and confusing them
is precisely the failure that ends an episode one cube short. The
PickXTimes episodes repeat the aliasing pattern of the simulation rollouts
on hardware: the odd-numbered panels are the same physical scene five times
over.

\subsection{Reproducibility Summary}
\label{sec:supp-repro}

Table~\ref{tab:real-repro} gathers the platform, control, training, and
verification settings needed to reproduce the physical experiments.

\begin{table}[htbp]
\centering
\caption{Reproducibility summary for the real-robot deployment.}
\label{tab:real-repro}
\small
\begin{tabular}{@{}lp{0.50\columnwidth}@{}}
\toprule
Component & Setting \\
\midrule
Control / camera rate & 30\,Hz, $640\times480$ RGB $\times 2$ \\
State / action        & 7-D (6 joints $+$ gripper width) \\
Action chunk          & 20 frames; replan every 10 frames or on subgoal
                        switch \\
Policy input images   & front $+$ wrist, letterboxed $224^2$ \\
Policy fine-tuning    & 40k steps, batch 128, lr $5\times10^{-5}$ after 10k
                        warmup, frozen encoder \\
Grasp verification    & lift $0.03$\,m (proprioceptive); $\delta_G = 8$\,px,
                        $6\times6$ grid, $60$\,px half-width \\
Placement head        & frozen SigLIP $+$ MLP, 4624-D input, $K = 4$ frames
                        per side \\
Placement thresholds  & target $0.5$; bin $0.05$ with in-bin release gate \\
Rejection bounds      & target 2, bin 3; grasp stuck timeout $2.7$\,s \\
Placement settling    & $0.4$\,s before verification \\
Safety                & $1.5$\,s ramp; $\Delta$joint $0.1$\,rad/frame;
                        button wait $20$\,s \\
Episode budget        & 240\,s, operator scene reset between episodes \\
Demonstrations        & 194 episodes / 237{,}517 frames (100 BinFill,
                        94 PickXTimes) \\
Verifier data         & 526 positive / 75 negative placement events (84\% hard);
                        30\% failure rate in the training mix \\
\bottomrule
\end{tabular}
\end{table}

Wording templates, state-machine thresholds, and verification constants are
shared between the annotation pipeline and the deployment runtime through
one configuration module, and the system refuses to run when a conditioning
string generated at deployment does not match the training vocabulary
verbatim.

\section{Extended Limitations}
\label{sec:supp-limitations}

\paragraph{The verification vocabulary is anchored to prehensile events.}
AGM localizes candidate transitions from gripper-load events, so its
achievement vocabulary covers exactly those subgoals whose completion is
bracketed by a grasp or a release. Pushing, sliding, articulated-object
manipulation, and tool use under a persistent grasp produce no compatible
event and fall through to the frozen policy. This is why AGM matches rather
than improves on the base policy for the two benchmark tasks whose subgoals
are not gripper-anchored. Widening the coverage calls for a broader family
of interaction detectors, not a different memory: the pointer is indifferent
to how events are localized.

\paragraph{The task is a sequence, not a graph.}
Memory here is an ordered subgoal sequence with a single pointer, which
suffices for repetitive and counting-based manipulation, where the structure
is genuinely linear. It cannot express branching plans, partially ordered
subgoals, or subgoals satisfiable out of order, and rollback reaches only
the associated grasp of the current repetition. Recoverability is likewise a
property of the subgoal type, fixed when the instruction is expanded rather
than read off the scene, so an object that becomes unreachable for reasons
the type label does not anticipate is retried until the rejection bound
forces an advance.

\paragraph{Bounded rejections trade one failure mode for another.}
The deadlock bounds ensure that a persistently rejecting verifier cannot
stall an episode, but they achieve this by advancing the pointer on an
unverified subgoal, converting a liveness failure into a possible progress
error. The bounds are small and irreversible placement is judged
permissively so that they are rarely reached, but what AGM guarantees is
that unverified writes are rare and bounded, not that they never happen.

\paragraph{Verification quality bounds task performance.}
Every write rests on a verification decision, so the framework inherits the
verifier's error profile. A false acceptance records progress that did not
occur; a false rejection spends budget repeating a completed subgoal. The
threshold asymmetry between recoverable and irreversible placement responds
to the asymmetry in these costs, but it manages the trade-off rather than
dissolving it.

\paragraph{Scope of the physical evaluation.}
The real-robot study covers two task families on one 6-DoF platform with a
single gripper geometry and object class. The framework transfers without
retuning across the two backbones tested, but we do not claim the calibrated
thresholds transfer across platforms; Appendix~\ref{sec:supp-runtime} documents the
recalibration procedure.

\end{document}